\documentclass[10pt,twocolumn,letterpaper]{article}

\usepackage{cvpr}
\usepackage{times}
\usepackage{epsfig}
\usepackage{graphicx}
\usepackage{amsmath}
\usepackage{amssymb}

\usepackage{graphicx} 
\usepackage{subfigure}

\usepackage{url}            
\usepackage{booktabs}       
\usepackage{amsfonts}       
\usepackage{nicefrac}       
\usepackage{microtype}      
\usepackage{color}

\usepackage{algorithm}
\usepackage{algorithmic}
\usepackage{amssymb}
\usepackage{amsthm}
\usepackage{amsmath,epsfig}
\usepackage{epstopdf}

\usepackage{enumitem}

\usepackage{multirow}
\usepackage{multicol}

\newcommand{\TODO}[1]{\textcolor{blue}{}\textcolor{blue}{#1}}

\newcommand{\D}[1]{\frac{\partial \mathcal{L}}{\partial #1 }}

\newcommand{\X}[1]{\mathbf{X}_{#1}}
\newcommand{\Y}[1]{\mathbf{Y}_{#1}}

\newcommand{\NX}[1]{\widehat{\mathbf{X}}_{#1}} 

\newcommand{\WM}[1]{\mathbf{G}_{#1}} 
\newcommand{\CM}[1]{\Sigma_{#1}} 

\newcommand{\HWM}[1]{\widehat{\mathbf{G}}_{#1}} 
\newcommand{\HCM}[1]{\widehat{\Sigma}_{#1}} 

\usepackage[pagebackref=true,breaklinks=true,letterpaper=true,colorlinks,bookmarks=false]{hyperref}

\cvprfinalcopy 
\def\cvprPaperID{5066} 

\begin{document}
	
	\title{An Investigation into the Stochasticity of Batch Whitening}
\author{Lei Huang$^{\dag}$ \quad Lei Zhao \quad Yi Zhou$^{\dag}$ \quad  Fan Zhu$^{\dag}$ \quad  Li Liu$^{\dag}$ \quad Ling Shao$^{\dag}$\\
	$^{\dag}$Inception Institute of Artificial Intelligence (IIAI), Abu Dhabi, UAE\\
		{\tt\small $\{$lei.huang, yi.zhou,  fan.zhu, li.liu, ling.shao$\}$@inceptioniai.org~~~bhneo@126.com}
}
	
	\maketitle
\begin{abstract}
    Batch Normalization (BN) is extensively employed in various network architectures by performing standardization within mini-batches.
     A full understanding of the process has been a central target in the deep learning communities. 
    Unlike existing works, which usually only analyze the standardization operation, this paper investigates the more general Batch Whitening (BW). Our work originates from the observation that while various whitening transformations equivalently improve the conditioning, they show significantly different behaviors in discriminative scenarios and training Generative Adversarial Networks (GANs). 
    We attribute this phenomenon to the stochasticity that BW introduces.
    We quantitatively  investigate the stochasticity of  different whitening transformations and show that it correlates well with the optimization behaviors during training. 
    We also investigate how stochasticity relates to the estimation of population statistics during inference.
    Based on our analysis, we provide a framework for designing and comparing  BW algorithms in different scenarios. 
    Our proposed BW algorithm improves the residual networks by a significant margin on ImageNet classification.
     Besides, we show that the stochasticity of  BW  can improve the GAN's performance with, however, the sacrifice of the training stability. 
\end{abstract}

\vspace{-0.13in}
\section{Introduction}
\label{sec_intro}
 Normalization techniques have been extensively used for learning algorithms during data preprocessing \cite{1998_NN_Yann, 2009_TR_Alex, 2015_CVPR_He}.
 It has been shown  that centering, scaling and decorrelating the inputs  speeds up the training \cite{1998_NN_Yann}.
 Furthermore, whitening the input that combines all above operations, improves the conditioning of the Hessian, making the gradient descent updates similar to the Newton updates \cite{1998_NN_Yann, 2011_NIPS_Wiesler, 2018_CVPR_Huang}.

 Batch Normalization (BN) \cite{2015_ICML_Ioffe} extends the idea of normalizing the input into the activations of intermediate layers of Deep Neural Networks (DNNs), and represents a milestone technique in the deep learning community \cite{2015_CVPR_He,2015_CoRR_Szegedy,2018_ECCV_Wu}.
 BN standardizes the activations by executing centering and scaling within a mini-batch of data, facilitating the optimization \cite{2015_ICML_Ioffe,2018_arxiv_Kohler,2018_NIPS_shibani} and  generalization \cite{2016_CoRR_Ba,2018_NIPS_Bjorck}. Batch Whitening (BW) further extends the scope of BN by removing the correlation of standardized activations \cite{2018_CVPR_Huang}. It has been shown to improve the performance in discriminative scenarios \cite{2018_CVPR_Huang} and Generative Adversarial Networks (GAN) \cite{2019_ICLR_Siaroin}. 
 
 \begin{figure}[tp]
 		\vspace{-0.16in}
 	\centering
 	\hspace{-0.2in}	\subfigure[BN standardization]{
 		\begin{minipage}[c]{.42\linewidth}
 			\centering
 			\includegraphics[width=3.6cm]{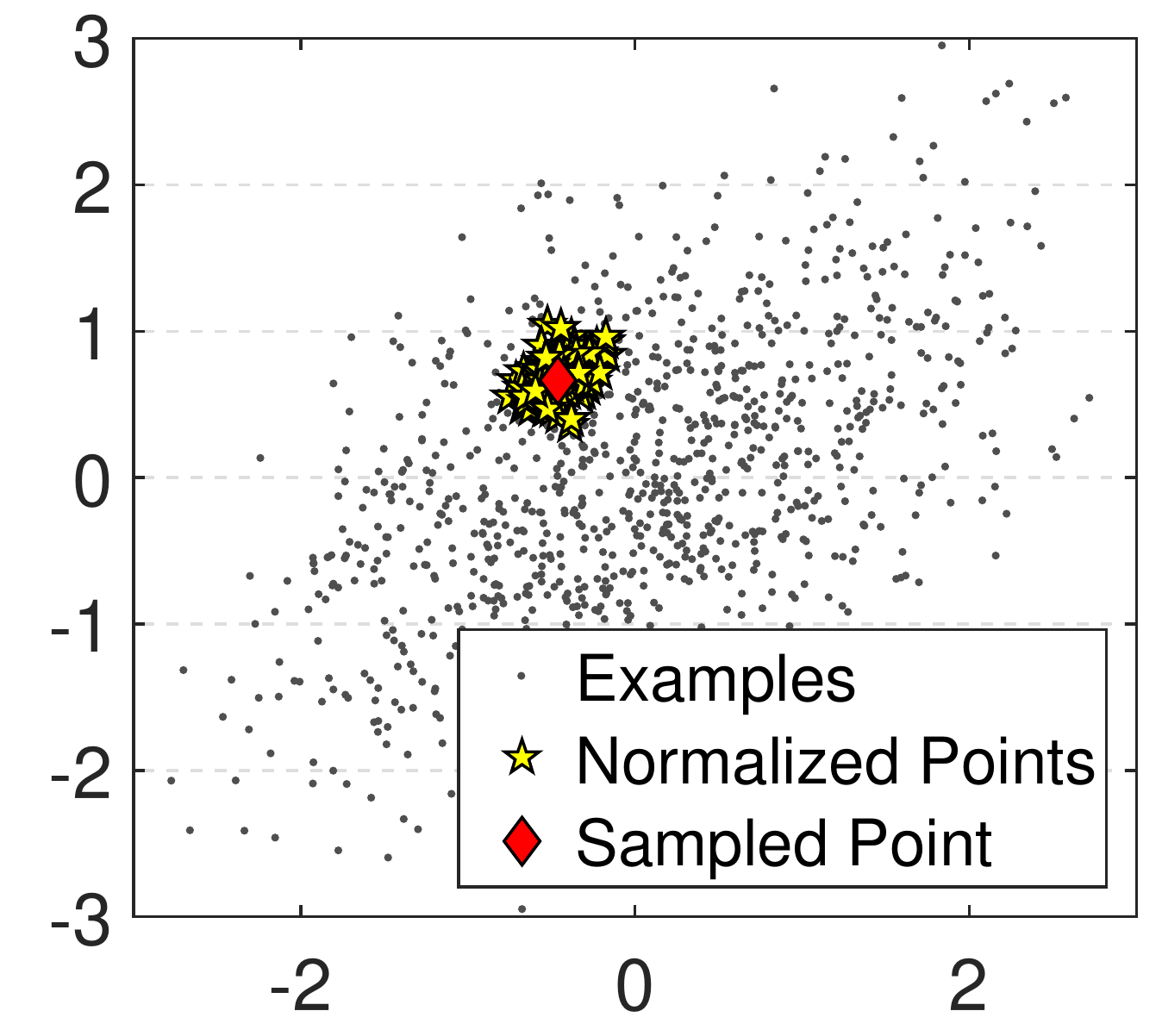}
 		\end{minipage}
 	}
 	\hspace{0in}	\subfigure[PCA whitening]{
 		\begin{minipage}[c]{.42\linewidth}
 			\centering
 			\includegraphics[width=3.6cm]{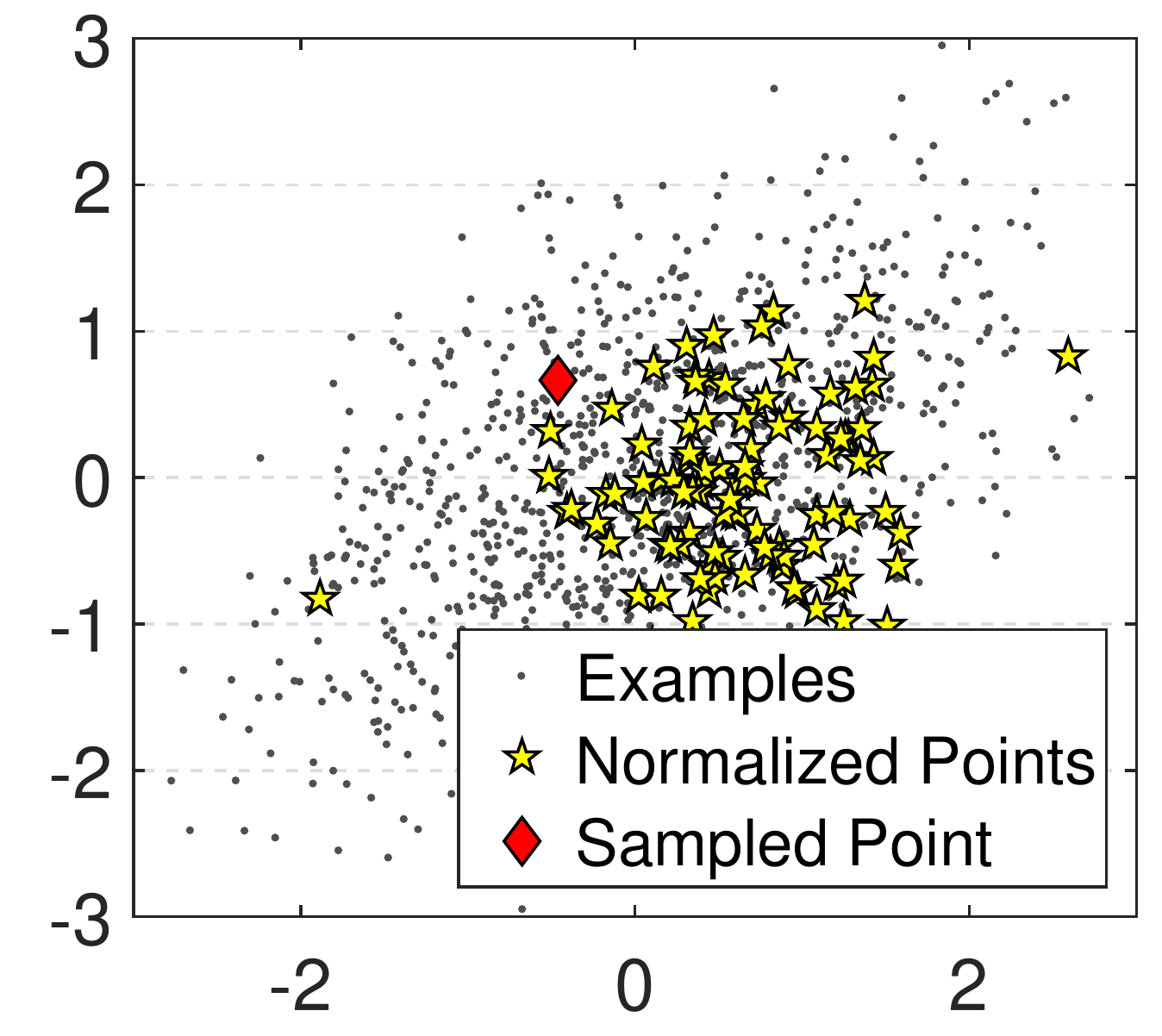}
 		\end{minipage}
 	}\\
 	\vspace{-0.14in}
 	\hspace{-0.2in}	\subfigure[ZCA whitening]{
 		\begin{minipage}[c]{.42\linewidth}
 			\centering
 			\includegraphics[width=3.6cm]{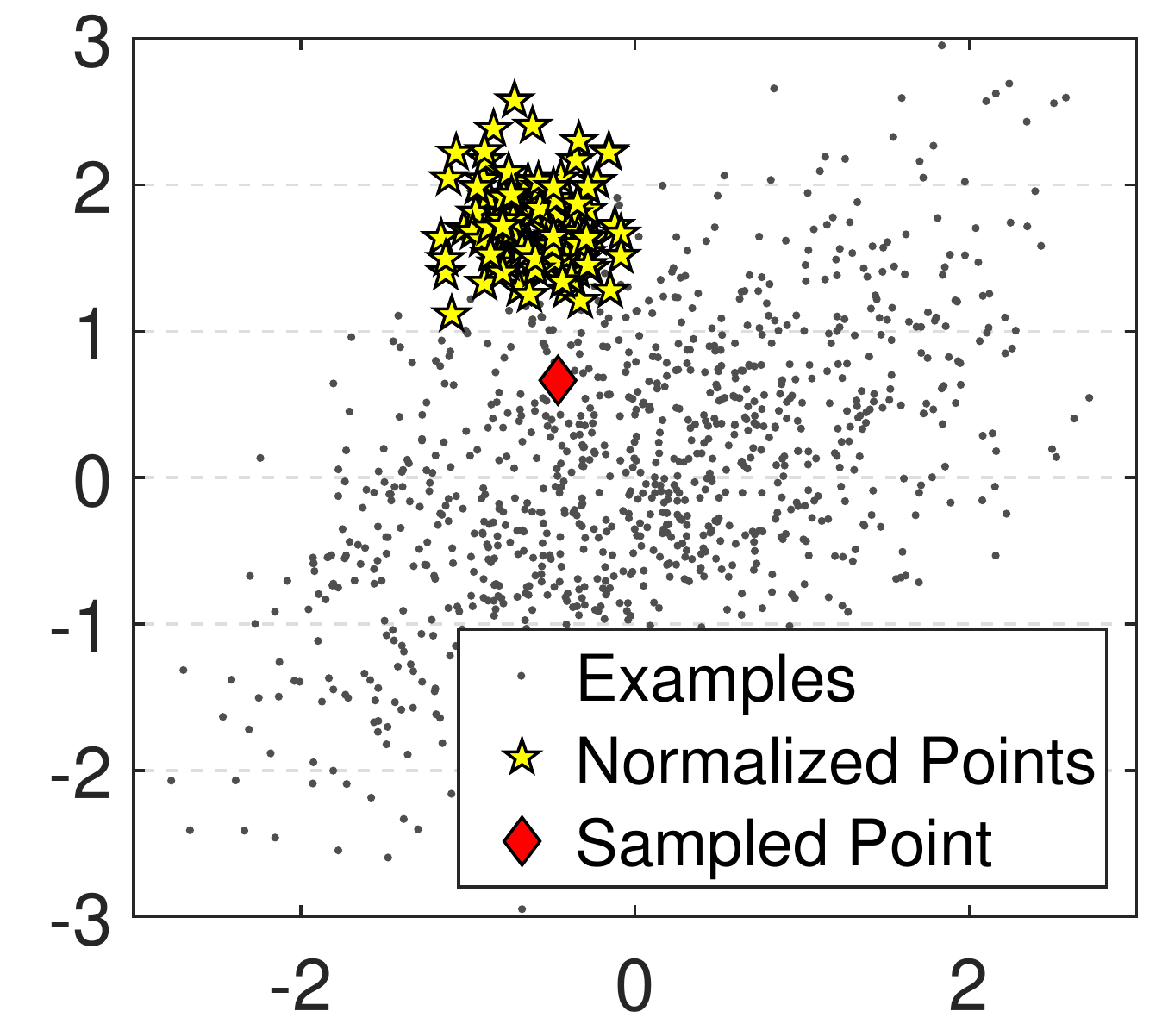}
 		\end{minipage}
 	}
 	\hspace{0in}	\subfigure[CD whitening]{
 		\begin{minipage}[c]{.42\linewidth}
 			\centering
 			\includegraphics[width=3.6cm]{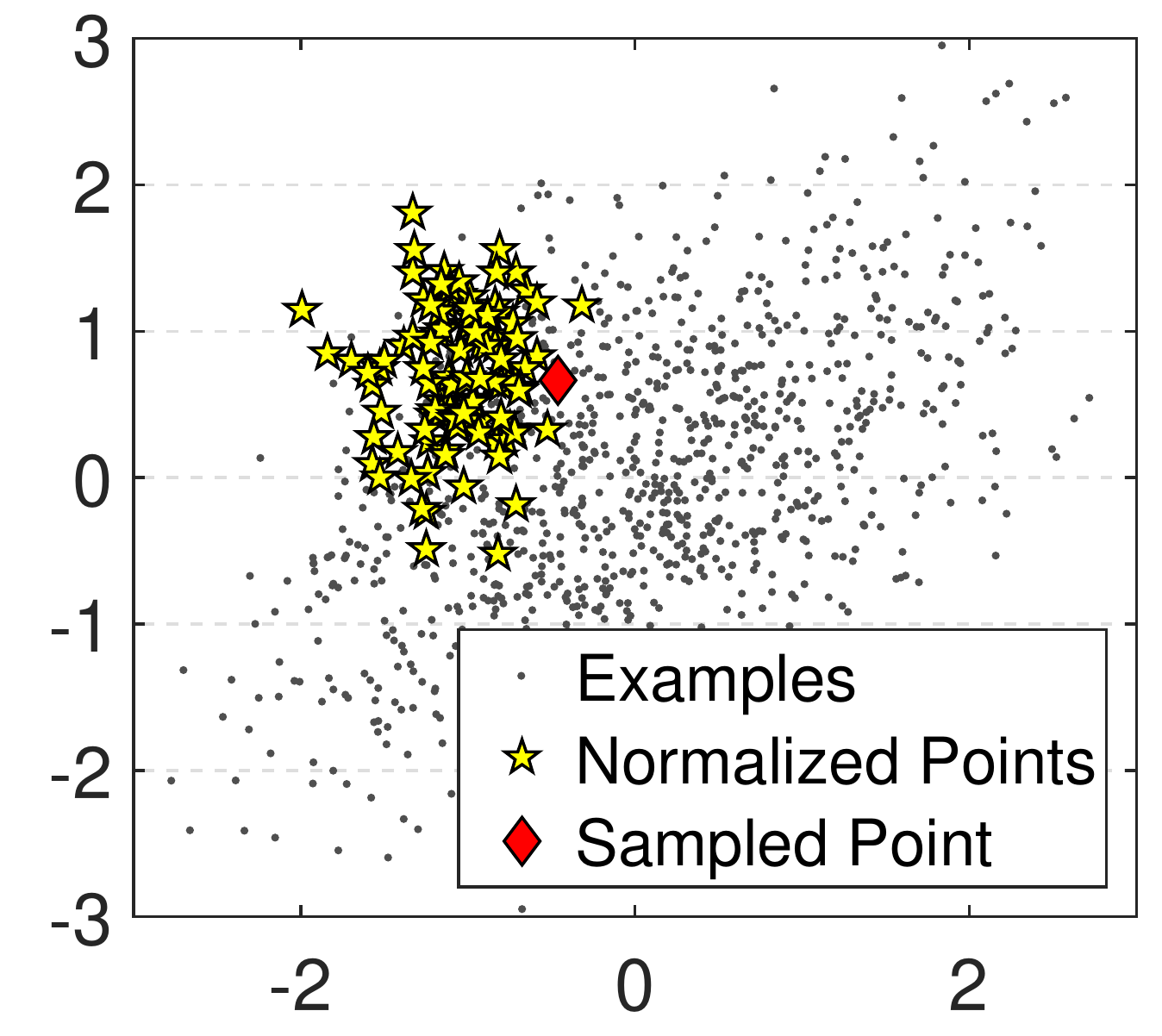}
 		\end{minipage}
 	}
 	\caption{ We sample 1000 examples (black points) from a Gaussian distribution in a 16-dimensional space, and show the examples in the two-dimension sub-space (the 6th and 16th dimension). Given an example $\mathbf{x}$ (red diamond), when combining with $100$ different mini-batches $\mathbf{X}^B$ ($B=64$),  we provide the normalized output $\hat{\mathbf{x}}$ (yellow pentagram), where (a), (b), (c) and (d) show the results of BN standardization, PCA, ZCA and CD whitening, respectively. }
 	\label{fig:Whitening_Stocha}
 	\vspace{-0.18in}
 \end{figure}

 Despite BW's theoretical support in improving conditioning, there remains some intriguing observations relating to BW that are not well explored. Firstly, while various whitening transformations can equivalently improve the conditioning \cite{2018_AS_Kessy}, they show significant differences in performance: 1)  Principal Component
 Analysis (PCA) whitening hardly converges while Zero-phase Component
 Analysis (ZCA) whitening works well in discriminative scenarios \cite{2018_CVPR_Huang}; 2) Cholesky Decomposition (CD) whitening achieves significantly better performance than ZCA in GAN training, while it has slightly worse performance in discriminative cases \cite{2019_ICLR_Siaroin}. Secondly, while group based whitening---where features are divided into groups and whitening is performed within each one---is essential in discriminative scenarios \cite{2018_CVPR_Huang,2019_ICCV_Pan}, full feature whitening has been shown to achieve better performance in training GANs \cite{2019_ICLR_Siaroin}. 
 
 This paper focuses on explaining the above observations of BW. 
 We find that the stochasticity introduced by normalization over batch data (Figure \ref{fig:Whitening_Stocha}) can be key to uncovering the  intriguing phenomena of BW. We quantitatively investigate the magnitude of stochasticity in different whitening transformations, using the evaluation method called Stochastic Normalization Disturbance (SND) \cite{2019_CVPR_Huang}.
  By doing so, we demonstrate that PCA whitening has significantly larger stochasticity, which is difficult to be controlled by either increasing batch size or using group based whitening. On the other side, ZCA whitening has the smallest stochasticity, while CD whitening has a moderate value, and more importantly, their stochasticity can be well controlled. This suggests ZCA whitening should have better optimization behaviors in training, while PCA whitening has problems in converging, due to the increased stochasticity which slows down   the progress of the optimization \cite{2014_JMLR_Nitish,2019_CVPR_Huang}. 
 
  We also investigate the stochasticity during inference, which is caused by the estimation of population statistics averaged over the mini-batch statistics during training. We show that in terms of estimating the population statistics of the whitening matrix, it is more stable to use the mini-batch covariance matrix indirectly (We calculate the whitening matrix after training) than the mini-batch whitening matrix directly. We further provide an empirical investigation to understand the reasons behind this observation, and find that the stochastic sequences of the mini-batch whitening matrix have a large diversity than the covariance matrix.

   Based on the above analyses, we provide a general framework for designing and comparing BW algorithms in different scenarios. We design new BW algorithm and apply them to the residual networks \cite{2015_CVPR_He} for ImageNet dataset  ~\cite{2009_ImageNet}, significantly improving the performance over the original one. 
 We further conduct thorough experiments on training GANs. We show that full feature whitening, when combined with coloring,  can improve the final score of evaluation. However, it reduces the training stability and is more sensitive to the hyper-parameters configurations. We attribute this phenomenon to two main effects caused by the introduced stochasticity of BW: 1) Strong stochasticity can increase the diversity of generated images and thus improve the GAN's performance; 2) At the same time, high stochasticity harms optimization and thus is more sensitive to the hyper-parameters. We argue that controlling the magnitude of whitening (stochasticity) is also important in training GANs and we validate this argument with our experiments. 

 \vspace{-0.06in}
\section{Related Work}
\label{sec_relatedWork}
\vspace{-0.06in}
The previous analyses on BN mainly focus on the optimization. One main argument is that BN can improve the conditioning of the optimization problem. This argument was initially introduced in the BN paper \cite{2015_ICML_Ioffe} and further refined in \cite{2018_NIPS_shibani},  showing that BN leads to  
 a smoother landscape of the optimization problem under certain assumptions. Ghorbani \etal \cite{2019_ICML_Ghorbani} investigated this explanation by computing the spectrum of the Hessian for a large-scale dataset. It is believed that the improved conditioning enables large learning rates, thus improving the generalization, as shown in \cite{2018_NIPS_Bjorck}.
 Another argument is that BN can adaptively adjust the learning rate \cite{2017_NIPS_Cho,2018_arxiv_Hoffer,2019_ICLR_Arora} due to its scale invariance \cite{2015_ICML_Ioffe,2016_CoRR_Ba}. This effect  has been  further discussed in combination with weight decay \cite{2019_ICLR_Zhang}. Other works have included an investigation into  the signal propagation and gradient back-propagation \cite{2019_ICLR_Yang}. Different from these approaches, our work focuses on  analyzing the stochasticity of  whitening over batch data. 
 
 The stochasticity introduced by normalization over batch data was first mentioned in the BN paper \cite{2015_ICML_Ioffe},  and further explored in \cite{2018_CoRR_Andrei,2018_ICML_Teye} from the perspective of Bayesian optimization. This stochasticity results in differences between the training distribution (using mini-batch statistics) and the test distribution (using estimated population statistics) \cite{2017_NIPS_Ioffe}, which is believed to be the main cause of the small-batch-problem of BN \cite{2018_ECCV_Wu}. To address this issue, a number of  approaches have been proposed \cite{2018_ECCV_Wu,2017_ICLR_Ren,2018_arxiv_luo,2017_NIPS_Ioffe,2018_NIPS_Wang,2019_ICCV_Singh}.
 Furthermore, it has been observed that BN also encounters difficulties in
 optimization during training \cite{2018_ACCV_Alexander,2019_CVPR_Huang}. This phenomenon is explored by the stochastic analysis shown in \cite{2019_CVPR_Huang}.  
 Different from the above research which focuses on standardization, we analyze, for the first time, the stochasticity on batch whitening. We propose that analyzing whitening rather than standardization, has several advantages in understanding the behaviors of normalization over batch data: 1) There are an infinite number of whitening transformations and the main ones show significant differences as discussed in Section \ref{sec_intro}; 2) The extent of the whitening (stochasticity) can be well controlled by the batch and group size, which provides more information in designing experiments. 
 
Our work is related to the previously proposed whitening methods regarding the activation of DNNs. One approach is to consider the whitening matrix as model parameters to be estimated using full data \cite{2015_NIPS_Desjardins,2017_ICML_Luo}. This kind of whitening has also been exploited in image style transformation tasks \cite{2017_NIPS_Li,2018_CVPR_Lu}. 
Another line of research is batch whitening, which is what this paper  discusses. This approach treats the normalization as a function over a mini-batch input. The main works include PCA whitening, ZCA whitening \cite{2018_CVPR_Huang} and its approximation ItN \cite{2019_CVPR_Huang},  and CD whitening \cite{2019_ICLR_Siaroin}. Pan \etal \cite{2019_ICCV_Pan} propose switchable whitening to learn different batch/instance whitening/standardization operations in DNNs. However, they used only the ZCA whitening transformation. Our work aims to understand different whitening transformation behaviors based on stochastic analysis.  

\vspace{-0.06in}
\section{Stochasticity Analysis of Batch Whitening}
\vspace{-0.05in}
\label{sec_method}
Let $\mathbf{X} \in \mathbf{R}^{d \times m}$ be a data matrix denoting the mini-batch input  of size $m$ in a certain layer. For simplifying the discussion, we assume that the data is centered, by performing $\X{} : = 	\mathbf{X} - \mathbf{\mu} \cdot \mathbf{1}^T$ where $\mathbf{\mu} = \frac{1}{m} \mathbf{X} \cdot \mathbf{1}$ is the mean of $\mathbf{X}$, and  $\mathbf{1}$ is a column vector of all ones.
 Whitening performs normalization
 over the mini-batch input as follows:
 \begin{small}
 	\setlength\abovedisplayskip{0.03in} 
 	\setlength\belowdisplayskip{0.03in}
	\begin{eqnarray}
	\label{eqn:whitening}
 \NX{} = \WM{} \X{},
	\end{eqnarray}
 \end{small}
\hspace{-0.02in}where $\WM{}$ is the mini-batch \textit{whitening matrix} that is derived from the corresponding \textit{covariance matrix} $\Sigma = \frac{1}{m} \mathbf{X}
 \mathbf{X}^T $.
 The population statistics of the  whitening matrix $\HWM{}$ used for inference,  is usually calculated by running average over the mini-batches as follows:
 \begin{small}
  	\setlength\abovedisplayskip{0.03in} 
  	\setlength\belowdisplayskip{0.03in}
 	\begin{eqnarray}
 	\label{eqn:running_average}
 	\HWM{} & = &(1-\lambda) \HWM{} + \lambda \WM{}.
 	\end{eqnarray}
 \end{small}
\hspace{-0.02in}It is clear that both the whitened output $\NX{}$ (Eqn. \ref{eqn:whitening}) and the population statistics $\HWM{}$ (Eqn. \ref{eqn:running_average}) can be viewed as stochastic variables,  because they  depend on the mini-batch inputs which are sampled over datasets. For illustration, we defer the analysis of the stochasticity to Sections \ref{sec:sto-training} and \ref{sec:sto-infer}, and first provide a review of the whitening transformations.

\vspace{-0.05in}
\subsection{Whitening Transformations}
\vspace{-0.05in}
There are an infinite number of possible whitening matrices, as shown in \cite{2018_AS_Kessy,2018_CVPR_Huang}, since any whitened data with a rotation is still whitened. This paper focuses on the whitening transformations based on PCA, ZCA and CD, since these three transformations have shown significant differences in performance when used in training DNNs \cite{2018_CVPR_Huang,2019_ICLR_Siaroin}.
Note that BN \cite{2015_ICML_Ioffe} can be viewed as a special case of batch whitening, since it performs standardization without removing correlations, where $\WM{BN}= (\mbox{diag}(\CM{}))^{-1/2}$ with $\mbox{diag}(\cdot)$ setting the off-diagonal elements of a matrix to zeros. 
To simplify the description, this paper regards BN as a (reduced) whitening transformation, unless otherwise stated.

\vspace{-0.18in}
\paragraph{PCA Whitening}
 uses $\WM{PCA}=\Lambda^{-\frac{1}{2}} \mathbf{D}^T$, where  $\Lambda=\mbox{diag}(\sigma_1, \ldots,\sigma_d)$ and $\mathbf{D}=[\mathbf{d}_1, ...,
\mathbf{d}_d]$ are the eigenvalues and associated eigenvectors of $\Sigma$, \ie $\Sigma = \mathbf{D}
\Lambda \mathbf{D}^T$. Under this transformation, the variables are first rotated by the eigen-matrix ($\mathbf{D}$) of the covariance,  then  scaled by the square root inverse of the eigenvalues ($\Lambda^{-\frac{1}{2}}$). PCA whitening over batch data suffers significant instability in training DNNs, and hardly converges, due to the so called Stochastic Axis Swapping (SAS), as explained in \cite{2018_CVPR_Huang}.

\vspace{-0.16in}
\paragraph{ZCA Whitening}
 uses $\WM{ZCA}=\mathbf{D} \Lambda^{-\frac{1}{2}} \mathbf{D}^T$, where the  PCA whitened input is rotated back by the corresponding rotation matrix $\mathbf{D}$. ZCA whitening works by stretching/squeezing  the dimensions along the eigenvectors. It has been shown that ZCA whitening avoids the SAS and achieves better performance over standardization (used in BN) on discriminative classification tasks  \cite{2018_CVPR_Huang}.

\vspace{-0.16in}
\paragraph{CD Whitening}
 uses $\WM{CD}=\mathbf{L}^{-1}$ where $\mathbf{L}$ is a lower triangular matrix from the Cholesky decomposition, with $\mathbf{L} \mathbf{L}^T=\CM{}$. This kind of whitening works by recursively decorrelating the current dimension over the previous decorrelated ones, resulting in a triangular form of its whitening matrix. CD whitening has been shown to achieve the state-of-the-art performance in training GANs, while ZCA whitening has degenerated performance. 

The primary motivation of this paper is to investigate the following problem: while all whitening methods equivalently improve the conditioning of the layer input, why do they show significantly different behaviors in training DNNs?
In the following sections, we provide a unified analysis and demonstrate that the key is the stochasticity introduced by normalization over batch data.
\begin{figure}[t]
	\centering
		\vspace{-0.15in}
	\hspace{-0.3in}	\subfigure[]{
		\begin{minipage}[c]{.44\linewidth}
			\centering
			\includegraphics[width=4.1cm]{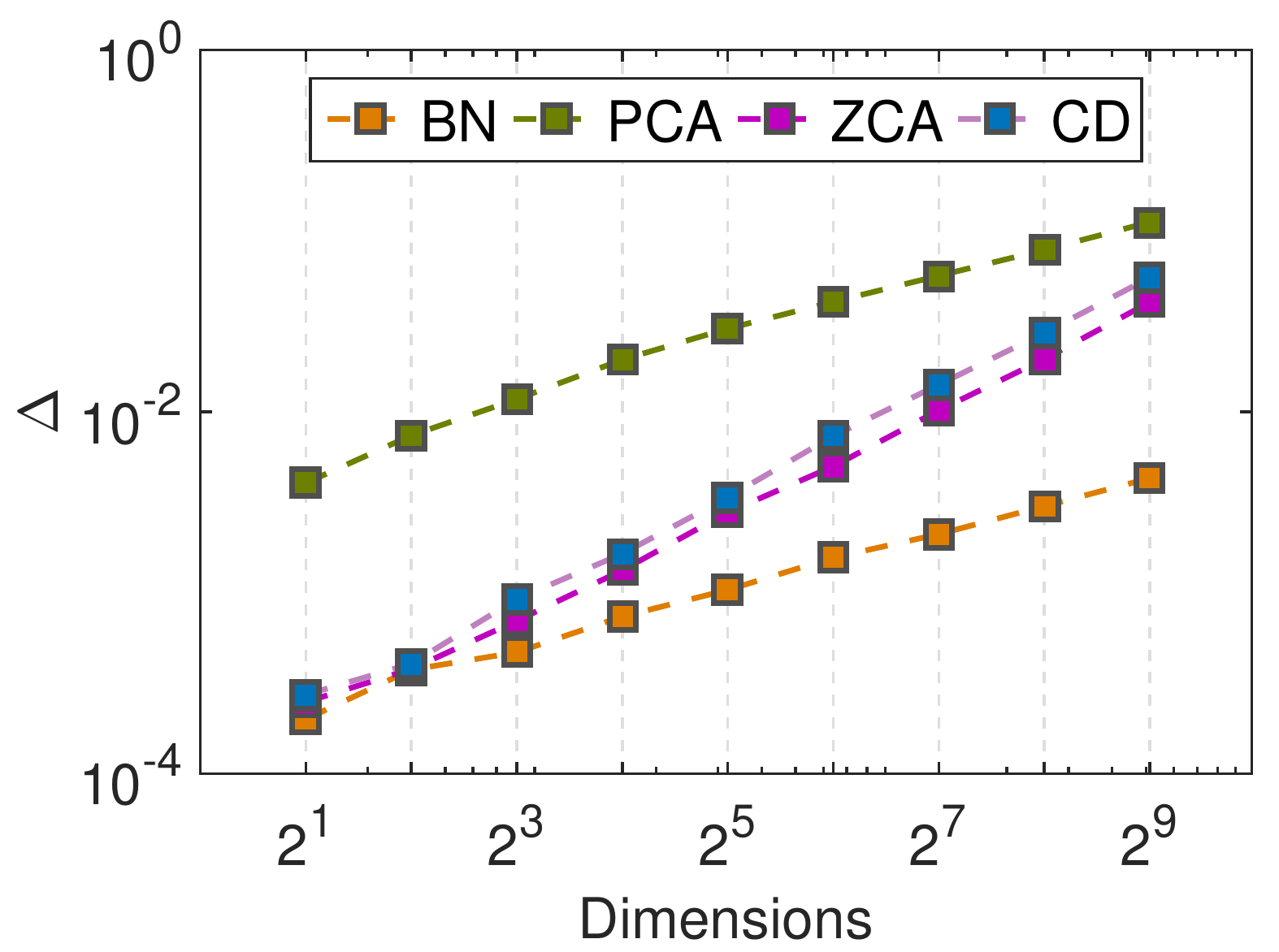}
		\end{minipage}
	}
	\hspace{0.1in}	\subfigure[]{
		\begin{minipage}[c]{.44\linewidth}
			\centering
			\includegraphics[width=4.1cm]{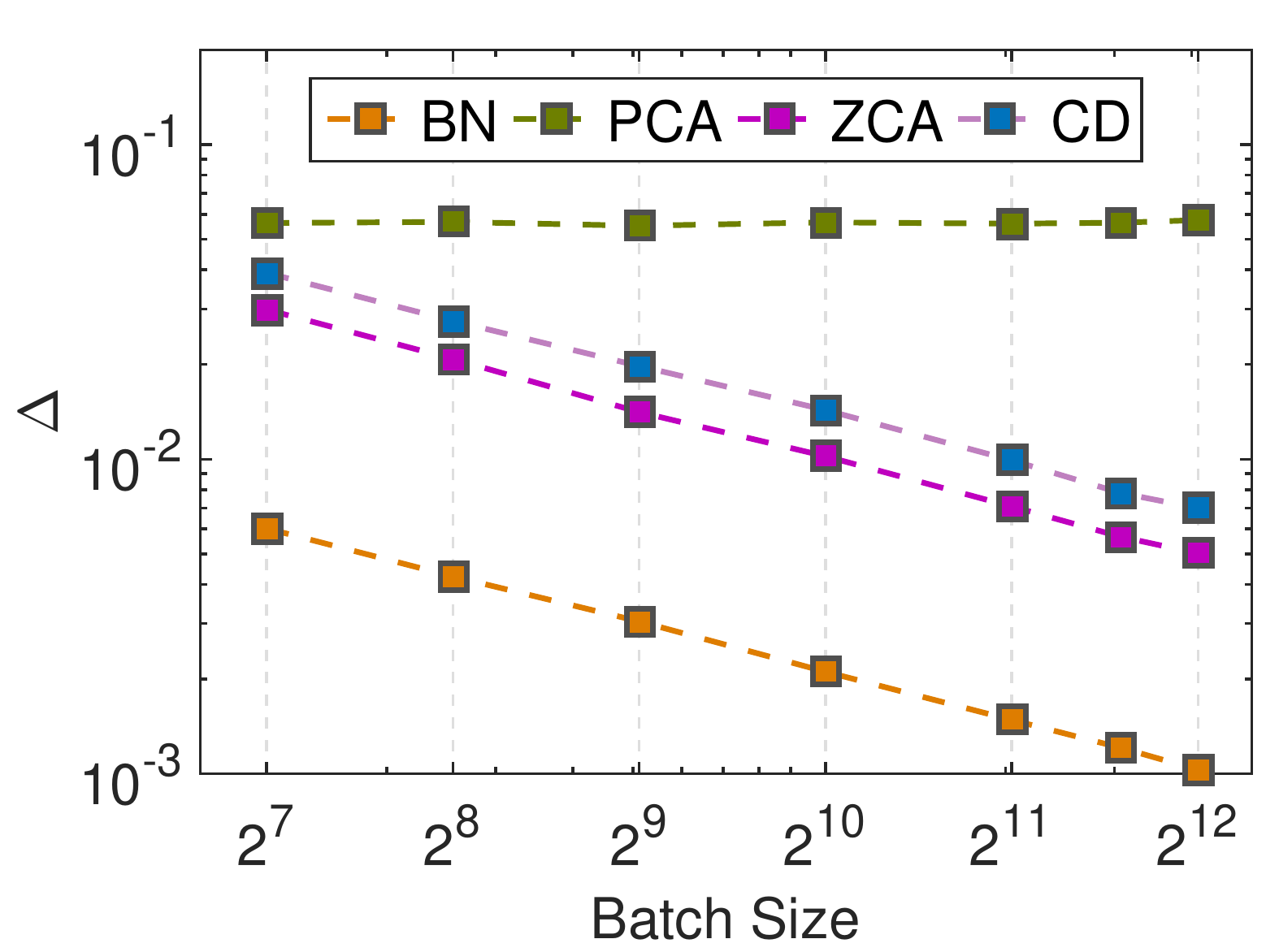}
		\end{minipage}
	}
	\vspace{-0.06in}
	\caption{SND comparison of different batch whitening methods.  We sample 60,000 examples from a Gaussian distribution as the training set. To calculate SND, we use $s=200$ and $N=20$.  We show (a) the SND with respect to the dimensions ranging from $2^1$ to $2^9$, under a batch size of 1024; (b) the SND with respect to the batch size ranging from $2^7$ to $2^{12}$, under a dimension of 128. }
	\label{fig:SND_analysis}
	\vspace{-0.2in}
\end{figure}

\vspace{-0.05in}
\subsection{Stochasticity During Training}
\vspace{-0.05in}
\label{sec:sto-training}



Given a sample $\mathbf{x} \in \mathbb{R}^d$ from a distribution $P_{\chi}$, we take a sample set $\mathbf{X}^B=\{\mathbf{x}_1,...,\mathbf{x}_B, \mathbf{x}_i \sim P_{\chi} \}$ with a size of $B$. The whitened output $\hat{\mathbf{x}}$ can be formulated as $\hat{\mathbf{x}}=\WM{}( \mathbf{X}^B ;\mathbf{x})$.
For a certain $\mathbf{x}$,  $\mathbf{X}^B$ can be viewed as a random variable \cite{2018_CoRR_Andrei,2018_ICML_Teye,2019_CVPR_Huang}. $\hat{\mathbf{x}}$ is thus another random variable showing stochasticity. Here, we investigate the stochasticity effects of different whitening transformations.

To provide a more intuitive illustration, we conduct a toy experiment to show how the normalized output of one sample  changes when combined with a different sample set $\mathbf{X}^B$, by performing different whitening methods. Figure \ref{fig:Whitening_Stocha} (a), (b), (c) and (d) show the results when performing BN, PCA, ZCA and CD whitening, respectively.
 It is clear that the distribution of the PCA whitened output of one sample is very sparse, which means that $\hat{\mathbf{x}}$ has significant diversity.
 This suggests that PCA whitening provides large stochasticity. On the other side, the BN standardized output shows a tight Gaussian-style distribution, which suggests that BN has smaller stochasticity.
 Note that BN can not guarantee that the normalized output has an identity covariance matrix, while other whitening methods can.
 Similarly, we also observe that ZCA whitening provides reduced stochasticity, compared to CD.  In fact, ZCA whitening has been shown to minimize the total squared distance between the original and whitened variables \cite{2018_AS_Kessy,2018_CVPR_Huang}.
 We conjecture that such a property results in ZCA whitening having smaller stochasticity than CD.
\begin{figure}[]
	\centering
		\vspace{-0.16in}
	\hspace{-0.2in}	\subfigure[]{
		\begin{minipage}[c]{.44\linewidth}
			\centering
			\includegraphics[width=4.2cm]{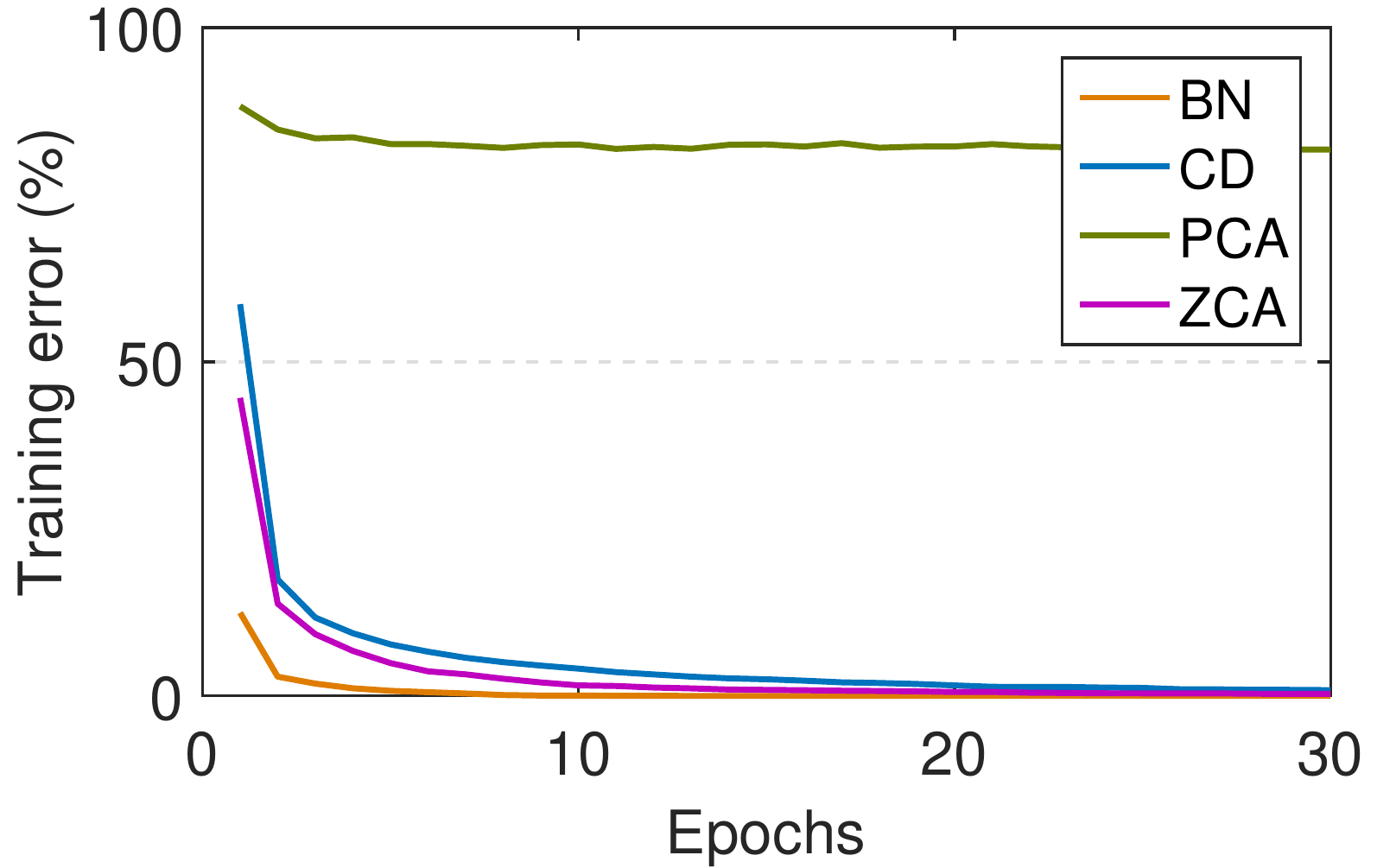}
		\end{minipage}
	}
	\hspace{0.1in}	\subfigure[]{
		\begin{minipage}[c]{.44\linewidth}
			\centering
			\includegraphics[width=4.2cm]{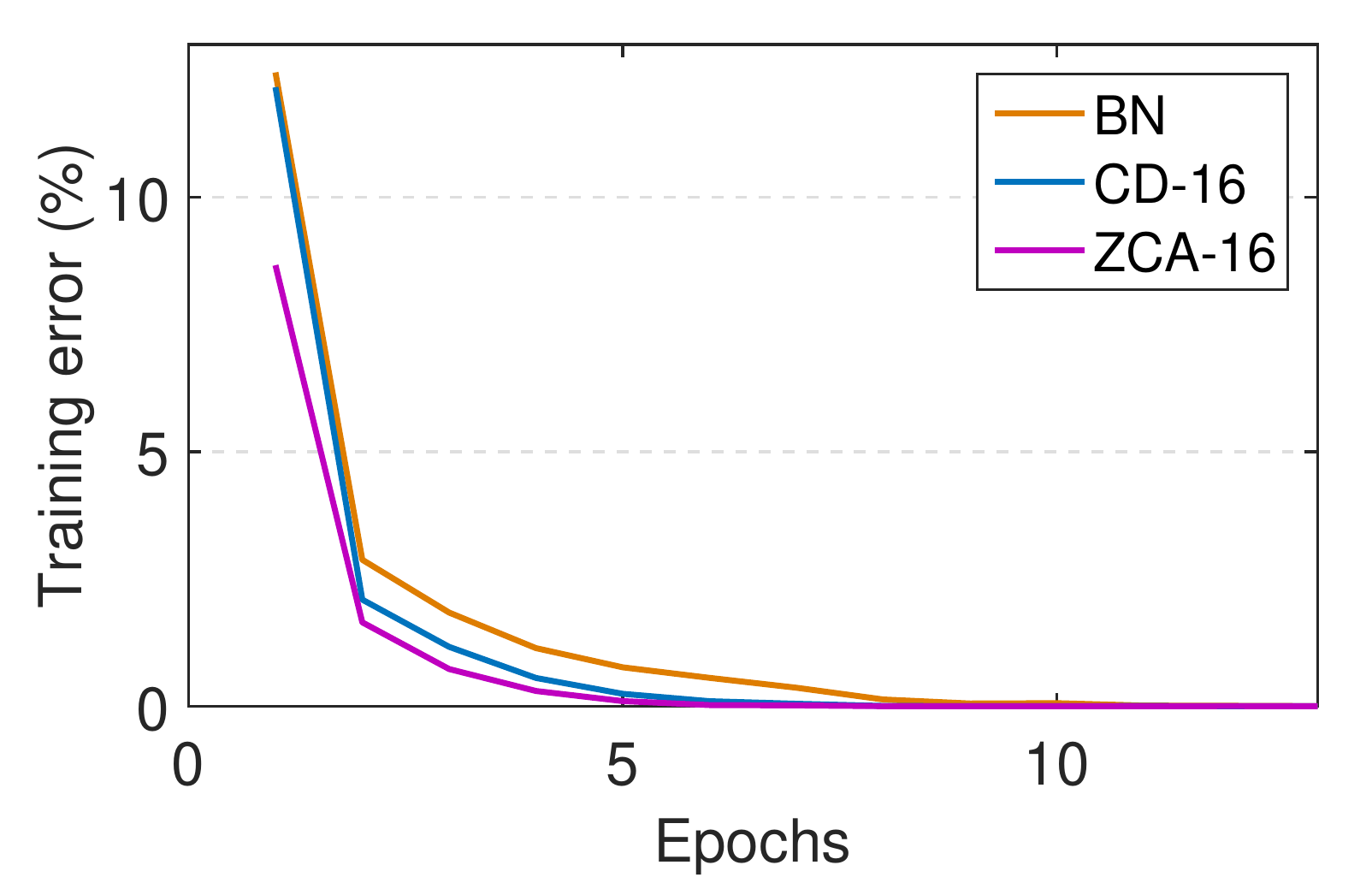}
		\end{minipage}
	}
		\vspace{-0.06in}
	\caption{ Experiments on a 4-layer MLP with 256 neurons in each layer, for MNIST classification. We use a batch size of 1024 and report the training errors. (a) The results of full whitening methods; (b) The results of group based whitening, where `ZCA-16' indicates ZCA whitening with a group size of 16. }
	\label{fig:MLP_train}
	\vspace{-0.20in}
\end{figure}
\vspace{-0.15in}
\subsubsection {Quantitative Analysis}
\vspace{-0.05in}
To provide a qualitative comparison, we exploit the evaluation called \emph{Stochastic Normalization Disturbance} (SND),  introduced in \cite{2019_CVPR_Huang}.
 The empirical estimation of SND for sample $\mathbf{x}$ over the normalization $\WM{}(\cdot)$ is defined as:
 \begin{small}
 	 	\setlength\abovedisplayskip{0.03in} 
 	 	\setlength\belowdisplayskip{0.03in}
 	\begin{eqnarray}
 	\label{eqn:empirical_SND}
       \widehat{\mathbf{\Delta}}_{\WM{}}(\mathbf{x})=\frac{1}{s} \sum_{i=1}^{s} \| \WM{}( \mathbf{X}^B_i ; \mathbf{x}) - \frac{1}{s} \sum_{j=1}^s \WM{}( \mathbf{X}^B_j ; \mathbf{x})     \|,
 	\end{eqnarray}
 \end{small}
\hspace{-0.05in}where $s$ denotes the number of mini-batches $\{\mathbf{X}^B_j\}_{j=1}^{s}$ that are randomly sampled from the dataset. SND can be used  to evaluate the stochasticity of  a sample after the normalization operation \cite{2019_CVPR_Huang}.
Further, a normalization operation $G(\cdot)$'s empirical SND is defined as  $\widehat{\mathbf{\Delta}}_{G}=\frac{1}{N}\sum_{i=1} ^{N} \widehat{\mathbf{\Delta}}(\mathbf{x}_i)$ given $N$ samples.  $\widehat{\mathbf{\Delta}}_{G}$ describes the magnitudes  of stochasticity for the corresponding normalization operations.

Here, we conduct experiments to quantitatively evaluate the effects of different normalization methods. Noticeably, the stochasticity is related to the batch size $m$ and the dimension $d$.
Figure \ref{fig:SND_analysis} (a) shows the SND of different normalization methods with respect to the dimensions, when fixing the batch size to 1024. We find that PCA whitening shows the largest SND while BN the smallest, over all the dimensions, which is consistent with the observation shown in Figure \ref{fig:Whitening_Stocha}. We notice that all whitening methods have an increased SND when the dimension increases. Besides, ZCA has a smaller SND than CD, over all the dimensions, which is also consistent with the data shown in Figure \ref{fig:Whitening_Stocha}.
 Figure \ref{fig:SND_analysis} (b) shows the SND of different normalization methods with respect to the batch size, when fixing the dimension to 128.
An interesting observation is that PCA whitening has nearly the same large SND among different batch sizes. This suggests that the PCA whitening is extremely unstable, no matter how accurate the estimation of the mini-batch covariance matrix is. This effect is in accordance with the explanation of Stochastic Axis Swapping (SAS) shown in \cite{2018_CVPR_Huang}, where a small change over the examples (when performing PCA whitening) results in a large change of representation.

To further investigate how this stochasticity affects DNN training, we perform experiments on a four-layer Multilayer Perceptron (MLP) with 256 neurons in each layer. We evaluate the training loss with respect to the epochs, and show the results in Figure \ref{fig:MLP_train} (a). We find that, among all the whitening methods, ZCA works the best, while PCA is the worst. We argue that this correlates closely with the SND they produce. Apparently, the increased stochasticity can slow
down training, even though all the whitening methods have equivalently improved conditioning.
 An interesting observation is that, in this case, BN works better than ZCA whitening. This is surprising since ZCA has improved conditioning over BN by removing the correlation \cite{2018_CVPR_Huang}, and it should theoretically have a better optimization behavior. However, the amplified stochasticity of ZCA whitening mitigates this advantage in optimization, thus resulting in a degenerated performance. Therefore,   from an  optimization perspective, we should control the extent of the stochasticity.

 \begin{figure}[t]
 	\centering
 		\vspace{-0.16in}
 	\hspace{-0.3in}	\subfigure[]{
 		\begin{minipage}[c]{.44\linewidth}
 			\centering
 			\includegraphics[width=4.2cm]{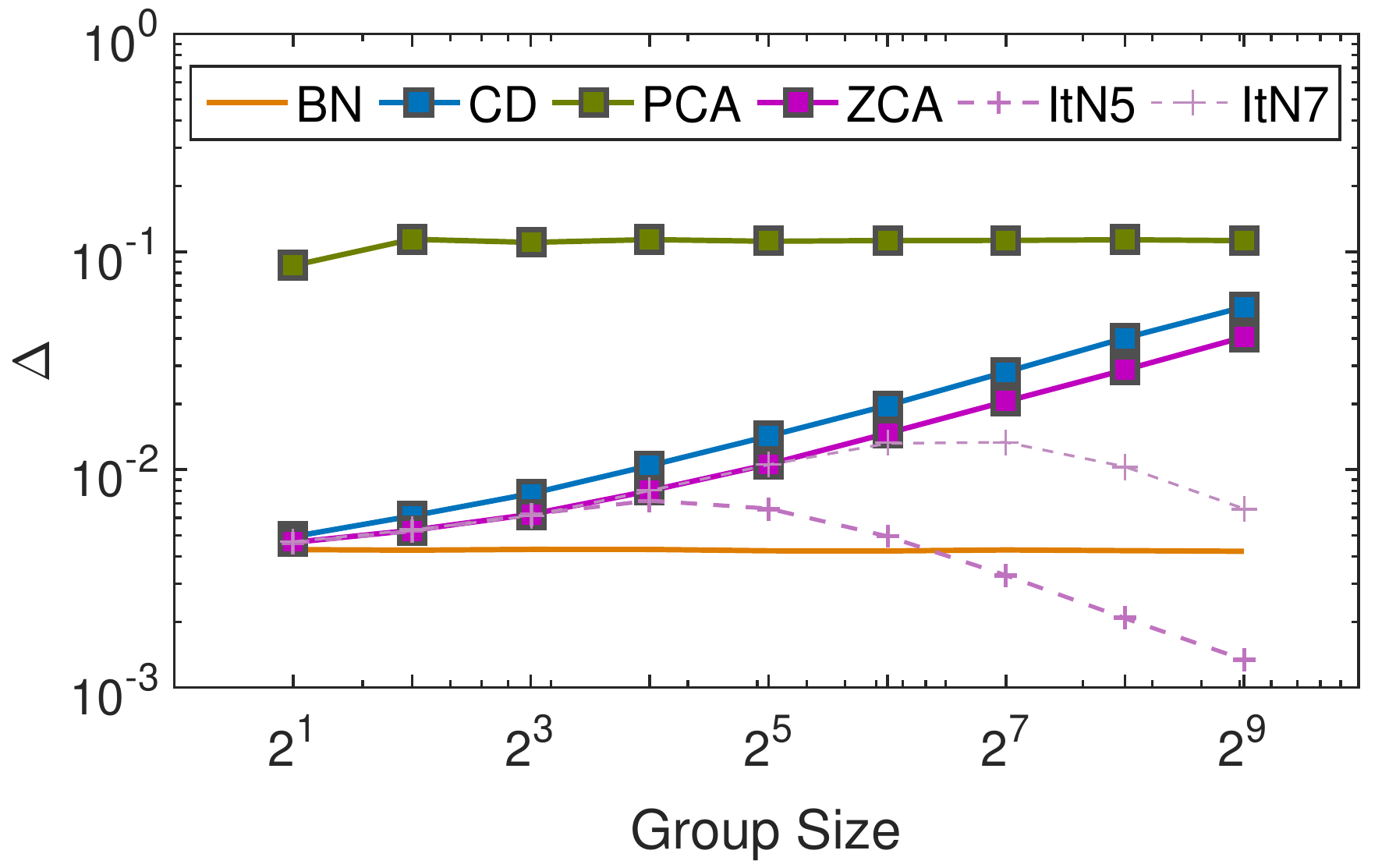}
 		\end{minipage}
 	}
 	\hspace{0.2in}	\subfigure[]{
 		\begin{minipage}[c]{.44\linewidth}
 			\centering
 			\includegraphics[width=4.2cm]{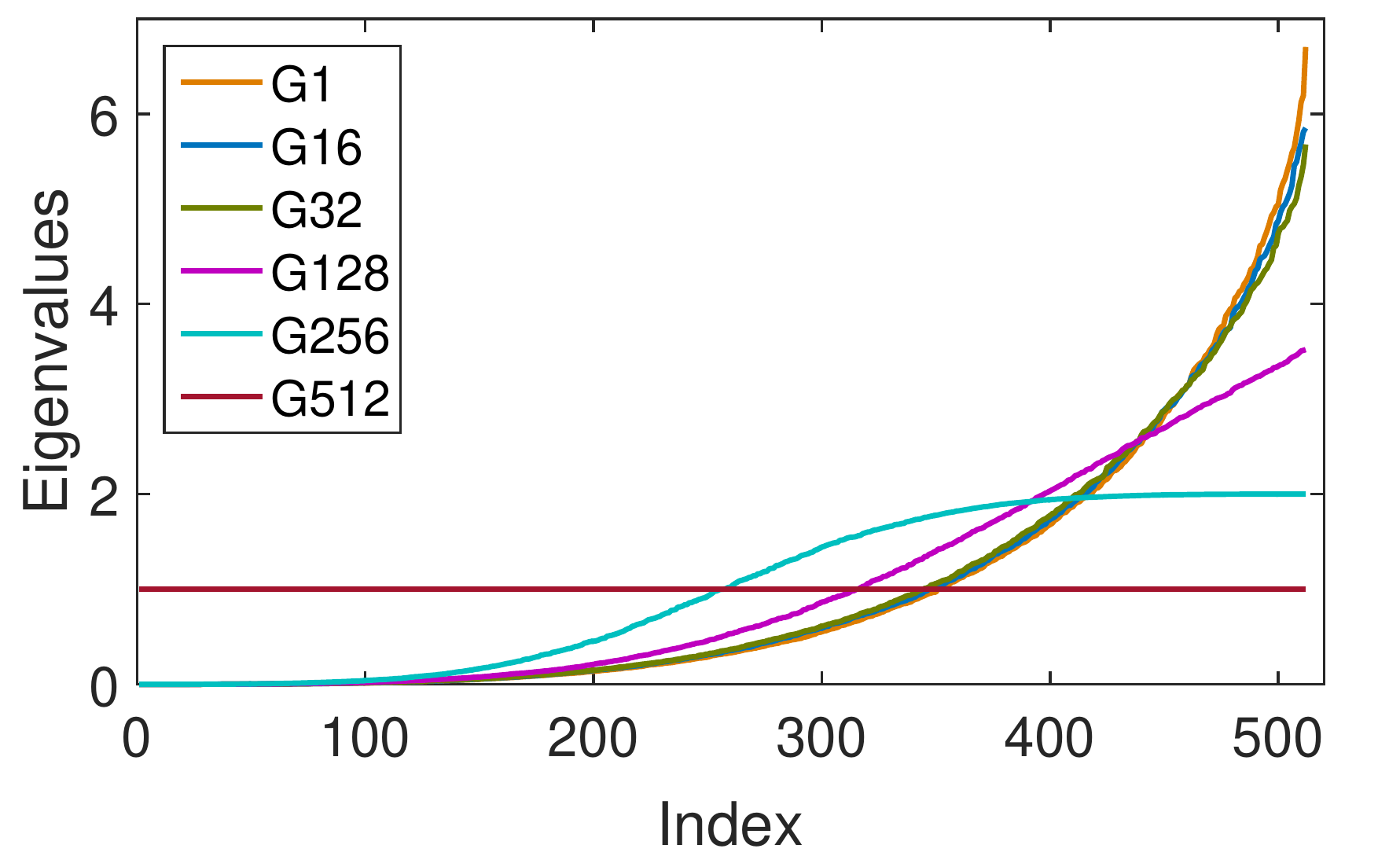}
 		\end{minipage}
 	}
 	\vspace{-0.06in}
 	\caption{Group-based whitening experiments. (a) We show the SND of different normalization operations with respect to the group size. The experimental setup is the same as Figure \ref{fig:SND_analysis} and the input dimension is $d=512$. (b) We show the spectrum of covariance matrix of the ZCA whitened  output (Note that CD/PCA whitening has the same spectrum as ZCA whitening.), where `G16' indicates whitening with a group size of 16. }
 	\label{fig:Group_SND}
 	\vspace{-0.20in}
 \end{figure}
 \vspace{-0.2in}
\subsubsection{Controlling the Stochasticity by Groups}
\vspace{-0.05in}
Huang \etal \cite{2018_CVPR_Huang} proposed  to use groups to control the extent of whitening.
They argue that this method reduces the inaccuracy in estimating the full covariance matrix when the batch size is not sufficiently large. Here, we empirically show how group based whitening affects the SND, providing a good trade-off between introduced stochasticity and improved conditioning. This is essential for achieving a better optimization behavior.

We evaluate the SND of different whitening transformations by varying the group size ranging from 2 to 512, as shown in Figure~\ref{fig:Group_SND} (a). We also display the spectrum of the covariance matrix of the whitened  output (based on groups) in Figure~\ref{fig:Group_SND} (b). We find that the group size effectively controls the SND of the ZCA/CD whitening.
With decreasing group size, ZCA and CD show reduced stochasticity  (Figure~\ref{fig:Group_SND} (a)), while having also degenerated conditioning (Figure~\ref{fig:Group_SND} (b)), since the output is only partially whitened.  Besides, we observe that  PCA whitening still has a  large SND over all group sizes, and with no significant differences. This observation further corroborates the explanation of SAS given in \cite{2018_CVPR_Huang}, \ie, that the PCA whitening is extremely unstable.

We also show the SND of the approximate ZCA whitening method (called ItN \cite{2019_CVPR_Huang}) in Figure~\ref{fig:Group_SND} (a), which uses Newton's iteration to approximately calculate the whitening matrix. We denote `ItN5' as the ItN method with an iteration number of 5.
An interesting observation is that ItN has smaller SND than BN, when using a large group size (\eg, 256) with a smaller iteration (\eg, T=5). This suggests that we can further combine group size and iteration number to control the stochasticity for ItN, providing an efficient and stable solution to approximate ZCA whitening \cite{2019_CVPR_Huang}.
The above observations furnish substantial insights into the application of whitening  in neural networks (especially in scenarios that requires decorating the activation in a certain layer), and we will further elaborate on this in Section~\ref{sec:experimental}.

We also use group-based ZCA/CD whitening methods on the four-layer MLP experiments. The results are shown in Figure \ref{fig:MLP_train} (b).  We observe that ZCA and CD whitening, with a group size of 16 to control the stochasticity, achieve better training behaviors than BN.

\begin{figure}[]
	\centering
		\vspace{-0.16in}
	\hspace{-0.15in}	\subfigure[ZCA Whitening]{
		\begin{minipage}[c]{.44\linewidth}
			\centering
			\includegraphics[width=4.0cm]{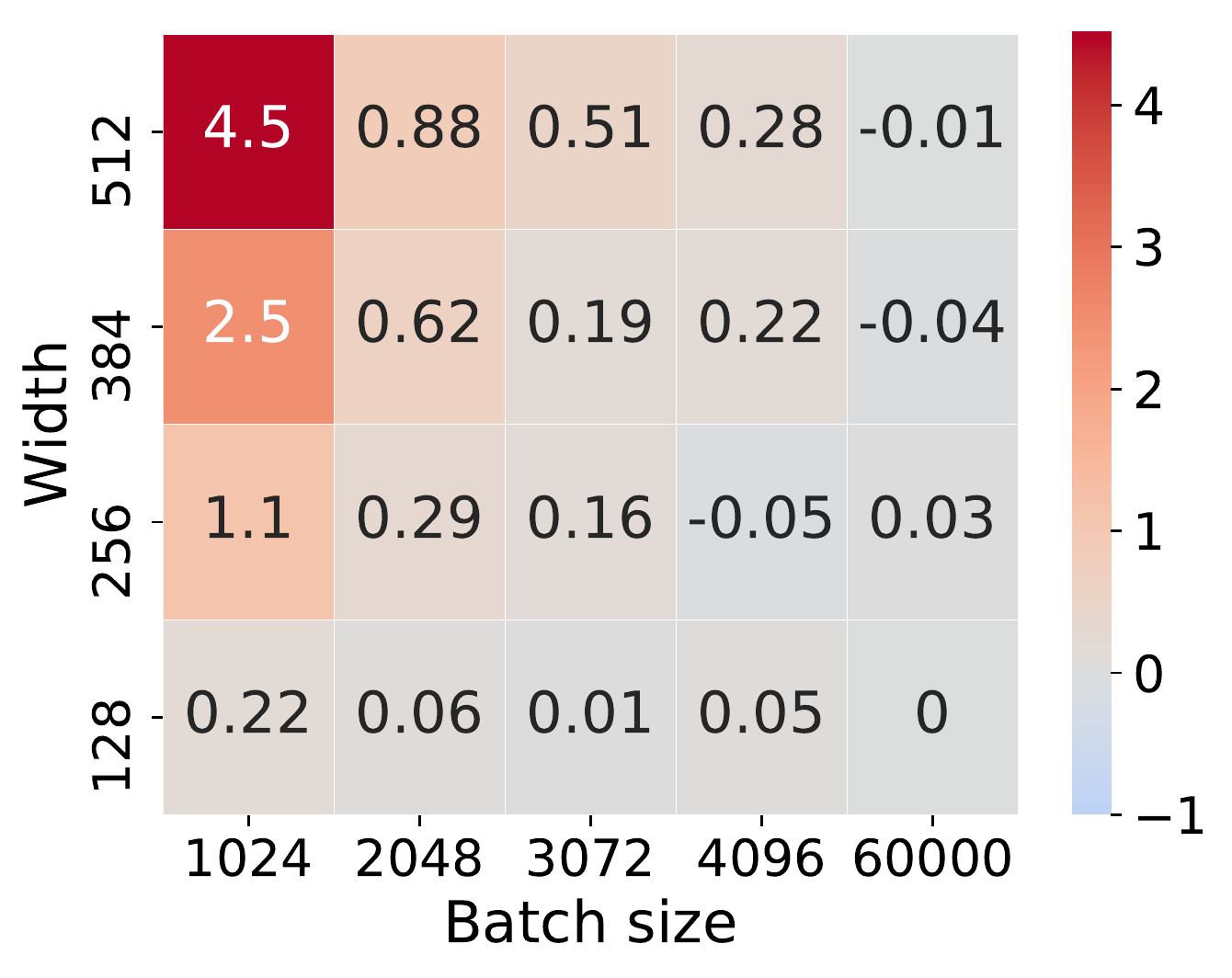}
		\end{minipage}
	}
	\hspace{0.2in}	\subfigure[CD Whitening]{
		\begin{minipage}[c]{.44\linewidth}
			\centering
			\includegraphics[width=4.0cm]{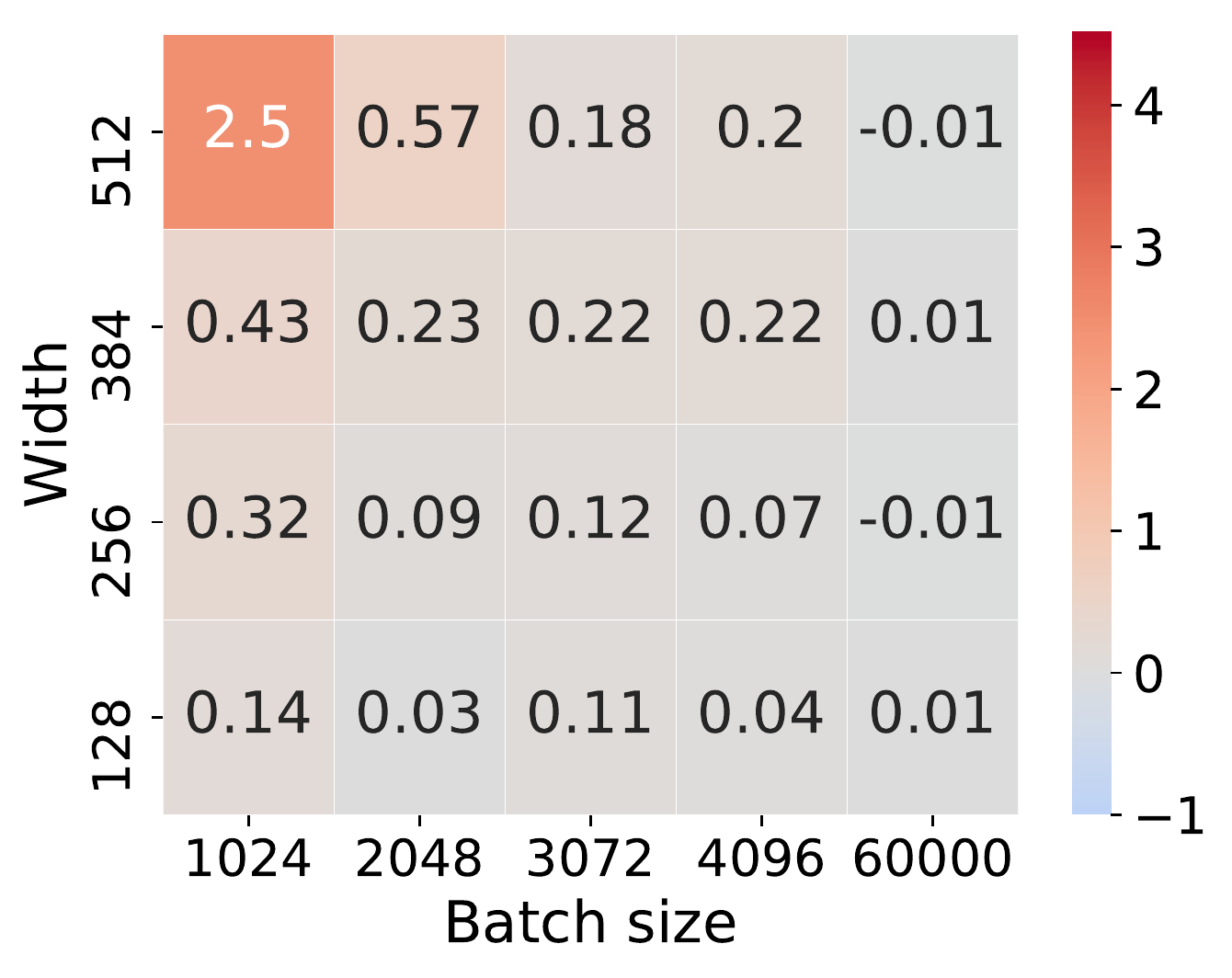}
		\end{minipage}
	}
		\vspace{-0.05in}
	\caption{Comparison of estimating $\HWM{}$ using different estimation objects ($\WM{}$/$\CM{}$). We train MLPs  with varying widths (number of neurons in each layer) and batch sizes, for MNIST classification.
		We evaluate the difference of test accuracy between  $\HWM{\CM{}}$ and $\HWM{\WM{}}$: $AC(\HWM{\CM{}})- AC(\HWM{\WM{}})$. (a) and (b) show the results using ZCA and CD whitening, respectively, under a learning rate of 1. We also tried other learning rates and obtained similar observations (See \TODO{Appendix}~\ref{sec-sup:trainingAndInfer} for details). }
	\label{fig:Estimation}
	\vspace{-0.20in}
\end{figure}

	\vspace{-0.05in}
\subsection{Stochasticity During Inference}
	\vspace{-0.05in}
\label{sec:sto-infer}
In the previous section we have shown that performing different whitening transformations,  by introducing different magnitudes of stochasticity, results in significantly different training behaviors from an optimization perspective. It's clear that such a stochasticity will also affect the final test performance during inference, because the population statistics $\HWM{}$ is estimated by running average (Eqn.~\ref{eqn:running_average}) over the stochastic sequences $\{ \WM{t} \}_{t=1}^{T}$, where T is the total number of training iterations.
The more diverse of the stochastic sequence  $\{ \WM{t} \}_{t=1}^{T}$, the more difficult to accurately estimate $\HWM{}$. Rather than estimating $\HWM{}$ directly, Siarohin \etal \cite{2019_ICLR_Siaroin} proposed to first estimate the population statistic of the covariance matrix $\HCM{}$, then compute $\HWM{}$ based on $\HCM{}$ after training.  However, no further analysis was provided to explain why they do like this.

Here, we provide an empirical investigation on how the estimation object ($\WM{}$/$\CM{}$) in Eqn.~\ref{eqn:running_average} affects the test performance. Let $\HWM{\WM{}}$ and $\HWM{\CM{}}$ denote the method to estimate $\HWM{}$ by $\WM{}$ and $\CM{}$, respectively.
 We conduct experiments on MLP with variations in width (the number of neurons in each layer) and batch size, for MNIST classification.
Figure~\ref{fig:Estimation} (a) and (b) show the results of ZCA and CD whitening, respectively.
We find that $\HWM{\CM{}}$ has better performance than  $\HWM{\WM{}}$, especially under the scenarios with large width and small batch size (Intuitively, estimating in high-dimensional space with a small batch size will make the estimation noisier).  This suggests that using $\CM{}$ to estimate  $\HWM{}$ indirectly  is more stable than using $\WM{}$ directly. Another interesting observation is that, by comparing Figure~\ref{fig:Estimation} (a) and (b), the differences between the estimation methods using CD whitening is smaller than the ZCA whitening.

\begin{figure}[]
	\centering
	\vspace{-0.16in}
	\hspace{-0.2in}	\subfigure[ histogram of $\delta(\HWM{\mathbf{M}}) $]{
		\begin{minipage}[c]{.44\linewidth}
			\centering
			\includegraphics[width=4.2cm]{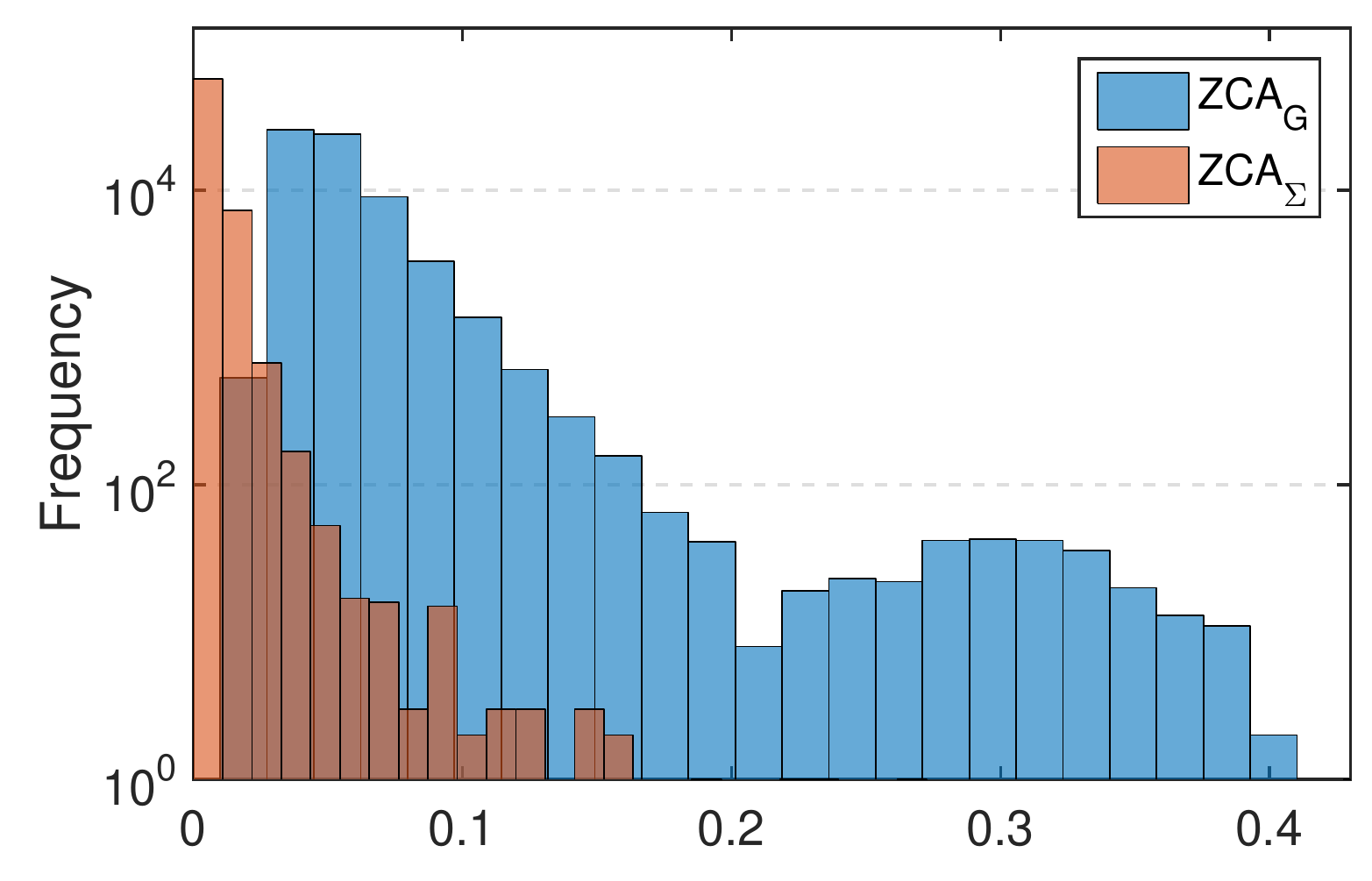}
		\end{minipage}
	}
	\hspace{0.15in}	\subfigure[ histogram of $\tilde{\delta} (\HWM{\mathbf{M}})$]{
		\begin{minipage}[c]{.44\linewidth}
			\centering
			\includegraphics[width=4.2cm]{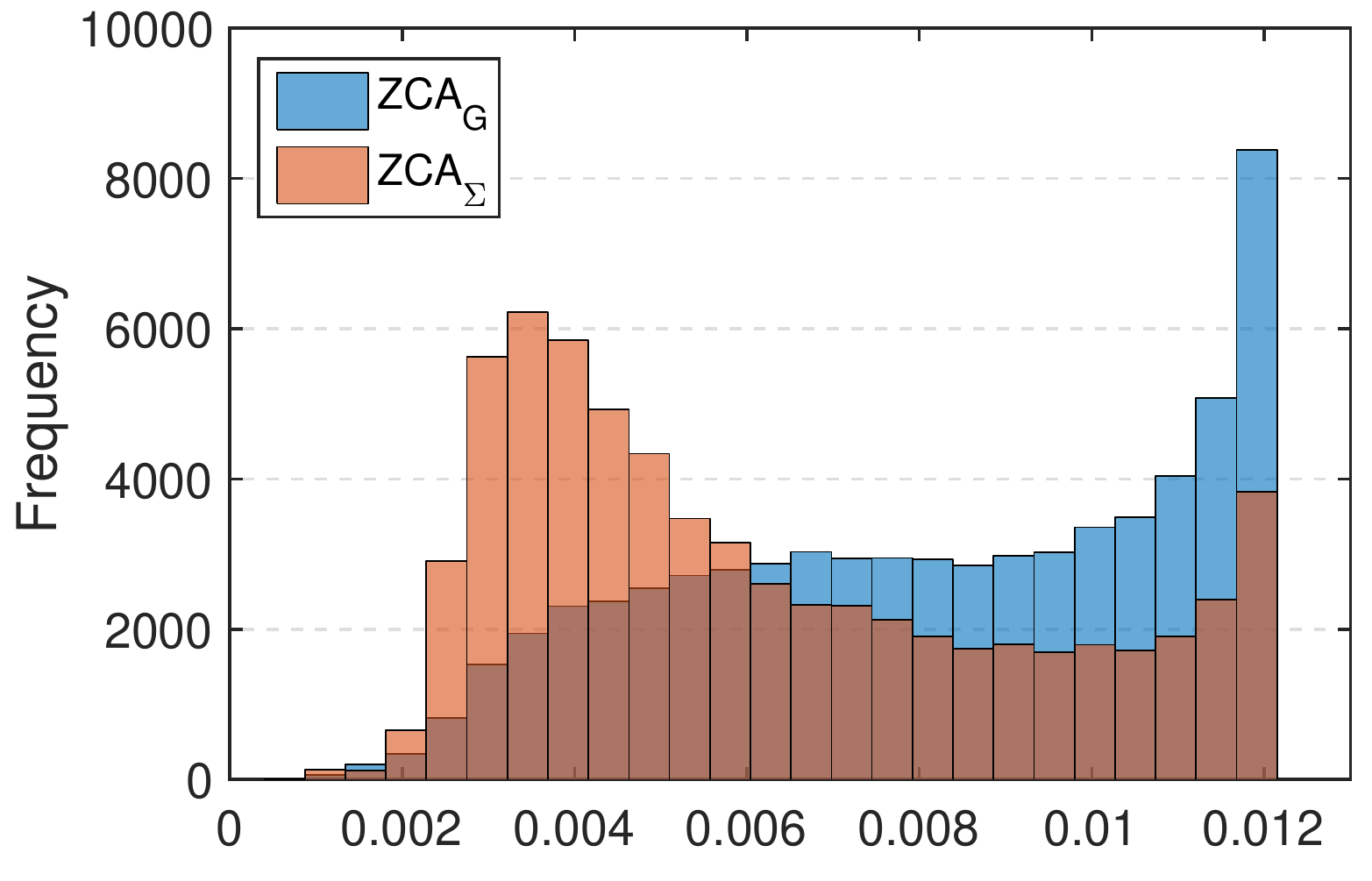}
		\end{minipage}
	}
		\vspace{-0.05in}
	\caption{Analysis of the diversity of the stochastic sequences in estimating $\HWM{}$. We report the histogram of $\delta(\HWM{\mathbf{M}}) $  and $\tilde{\delta} (\HWM{\mathbf{M}})$.}
	\label{fig:Histogram_std}
	\vspace{-0.20in}
\end{figure}

We further analyze how the diversity of stochastic sequences $\{\mathbf{M}_t\}_{t=1}^T$  affect the  estimation of $\HWM{}$, where $\mathbf{M} \in \{ \WM{},  \CM{} \}$.
Intuitively, if the stochastic sequence has high diversity, the effects of estimation will be worse.
 We view each element of $\HWM{}$ as an independent stochastic variable and calculate the standard deviation of each element during training, as follows:
  \begin{small}
  	 	\setlength\abovedisplayskip{0.05in} 
  	 	\setlength\belowdisplayskip{0.05in}
  	\begin{eqnarray}
  	\label{eqn:STD}
    \delta(\HWM{\mathbf{M}}^{ij}) = \sqrt{\frac{1}{T} \sum_{t=1}^{T} (\mathbf{M}_{t}^{ij} - \frac{1}{T}\sum_{t'=1}^{T}(\mathbf{M}_{t'}^{ij}) )^2 },
  	\end{eqnarray}
  \end{small}
\hspace{-0.05in}where $\HWM{}^{ij}$ ($\mathbf{M}_{t}^{ij}$)  indicates the (i,j)-th element of $\HWM{}$ ($\mathbf{M}_{t}$).
Furthermore, we calculate the normalized standard deviation of each element $\tilde{\delta} (\HWM{\mathbf{M}}^{ij})$, which, as defined in Eqn.~\ref{eqn:STD}, is calculated over  $\widetilde{\mathbf{M}}_t^{ij}=\mathbf{M}_t^{ij}/ \sqrt{\sum_{t=1}^T (\mathbf{M}_t^{ij})^2} $. Figure \ref{fig:Histogram_std} (a) and (b) show the histogram of $\delta(\HWM{\mathbf{M}}) $  and $\tilde{\delta} (\HWM{\mathbf{M}})$, respectively, when using ZCA whitening. We clearly find that $\HWM{\WM{}}$ has a larger standard deviation on average, thus a large diversity in general, compared to $\HWM{\CM{}}$. This reveals why using $\CM{}$ is more stable for estimating  $\HWM{\WM{}}$ than using $\WM{}$.

\vspace{-0.05in}
\section{Evaluation on Vision Tasks}
	\vspace{-0.05in}
\label{sec:experimental}
Based on the previous analysis, we can design new BW algorithms, and construct more effective DNNs by using BW. 
We investigate this in classification and training GANs. The code to reproduce the experiments is available at \textcolor[rgb]{0.33,0.33,1.00}{https://github.com/huangleiBuaa/StochasticityBW}.
	\vspace{-0.05in}
\subsection{The Landscape of Batch Whitening Algorithms}
\label{sec:experimental-landscape}
	\vspace{-0.05in}
We provide a general view of batch whitening algorithms in Algorithm~\ref{alg_forward} for vectorial inputs $\mathbf{X} \in \mathbb{R}^{ d \times m}$. Back-propagation is necessary to pass through the whitening transformation, and we provide the details in \TODO{Appendix}~\ref{sec-sup-BP} for completeness.   For feature map inputs $\mathbf{X}_F \in \mathbb{R}^{h \times w \times d
	\times m} $, where $h$ and $w$ indicate the height and width, the whitening transformation is performed  over the unrolled $\mathbf{X}
\in \mathbb{R}^{ d \times (m h w)}$,
 since each spatial position of the
feature map can be viewed as a sample \cite{2015_ICML_Ioffe}.

Note that an extra step to recover the representation capacity of normalization is given in Line 11 of Algorithm \ref{alg_forward}, which is shown to be empirically effective in \cite{2015_ICML_Ioffe,2016_CoRR_Ba,2018_CVPR_Huang,2018_ECCV_Wu}.
There are two alternatives for $\phi_3$: One is a dimension-wise scale and shift \cite{2015_ICML_Ioffe,2018_CVPR_Huang}: $\mathbf{Y}_k = \gamma_k \NX{k} + \beta_k, (k=1,...,d)$; The other is a  coloring transformation: $\mathbf{Y}=\mathbf{W} \NX{} + \mathbf{b}$, which was proposed in \cite{2019_ICLR_Siaroin} to achieve better performance in training GANs.
By combining whitening transformation $\phi_1$, estimation object $\phi_2$ and recovery operation $\phi_3$, we can design different BW algorithms. Table \ref{table:Alogorithm} shows the scope of value for different components this paper discusses. Note that ItN \cite{2019_CVPR_Huang} is the efficient and numerical stable approximation of ZCA whitening. We use the `Scale $\&$ Shift' operation in $\phi_3$ for all algorithm, unless otherwise stated.

\begin{algorithm}[tb]
	\algsetup{linenosize=\footnotesize}
	\footnotesize
	\caption{A general view of batch whitening algorithms.}
	\label{alg_forward}
	\begin{algorithmic}[1]
		\begin{small}
			\STATE \textbf{Input}: mini-batch inputs $ \mathbf{X} \in \mathbb{R}^{d \times m} $.
			\STATE \textbf{Output}: $ \mathbf{Y} \in \mathbb{R}^{d \times m}$.
			\IF {Training}
			\STATE	Calculate covariance matrix: $\CM{} = \frac{1}{m}\mathbf{X} \mathbf{X} ^T + \epsilon \mathbf{I}$.
			\STATE  Calculate whitening matrix: $\WM{} = \phi_1(\CM{})$.
			\STATE  Calculate whitened output: $\widehat{\mathbf{X}}= \WM{} \mathbf{X}$.
			\STATE  Update population statistics: $\HWM{} = \phi_2(\CM{}/\WM{})$.
			\ELSE
			\STATE  Calculate whitened output: $\NX{}= \HWM{} \mathbf{X}$.
			\ENDIF
			\STATE  Recover representation: $\mathbf{Y} = \phi_3(\NX{})$.
			%
		\end{small}
	\end{algorithmic}
\end{algorithm}

\begin{table}[t]
	\centering
	\begin{small}
		\begin{tabular}{ll}
		\bottomrule[1pt]
			COMPONENT    & VALUE \\
			\hline
			Whitening transformation & $\{$`ZCA', `PCA', `CD', `ItN' $\}$ \\
			Estimation object & $\{$ `$\CM{}$', `$\WM{}$' $\}$  \\
			Recovery operation & $\{$ `Scale $\&$ Shift', `Coloring' $\}$  \\
		\toprule[1pt]
		\end{tabular}
		\caption{The scope of value in Algorithm \ref{alg_forward} for different components this paper discusses. The Cartesian product of the values considers the landscape of the batch whitening algorithms used in this study.  }
		\label{table:Alogorithm}
	\end{small}
	\vspace{-0.20in}
\end{table}

	\vspace{-0.05in}
\subsection{Investigation on Discriminative Scenarios}
\vspace{-0.05in}
In the section, we first investigate how different whitening transformations affect the training of VGG network  \cite{2014_CoRR_Simonyan} for CIFAR-10 datasets  ~\cite{2009_TR_Alex}, from the  perspective of optimization.
We then show the performance on large-scale ImageNet classification  ~\cite{2009_ImageNet}, by applying the designed BW algorithm on residual networks \cite{2015_CVPR_He}.


	\vspace{-0.15in}
\subsubsection{VGG on CIFAR-10}
	\vspace{-0.05in}
\label{sec:VGG}
We use the VGG networks \cite{2014_CoRR_Simonyan} tailored for $32\times 32$ inputs (16 convolutional layers and 1 fully-connected layer).
We add  the normalization methods after each convolutional layer in the VGGs.
We compare several methods including: 1) The full whitening methods `PCA', `ZCA' and `CD'; 2) The approximate whitening `ItN' and the standardization `BN'; 3) Group-based whitening methods with the group size ranging in $\{512, 256, 128, 64, 32, 16\}$. We denote `ZCA-16' as ZCA whitening with group size 16. We focus on comparing the training performance from an optimization perspective, and we use mini-batch covariance matrix `$\CM{}$' as the estimation object for all methods during inference. 
We use SGD with a batch size of 256 to optimize the model. We set the initial learning rate to 0.1, then divide it by 5 after  60 epochs and finish the training at 120 epochs.

\begin{figure}[]
	\centering
	\vspace{-0.12in}
	\hspace{-0.2in}	\subfigure[Training error]{
		\begin{minipage}[c]{.44\linewidth}
			\centering
			\includegraphics[width=4.2cm]{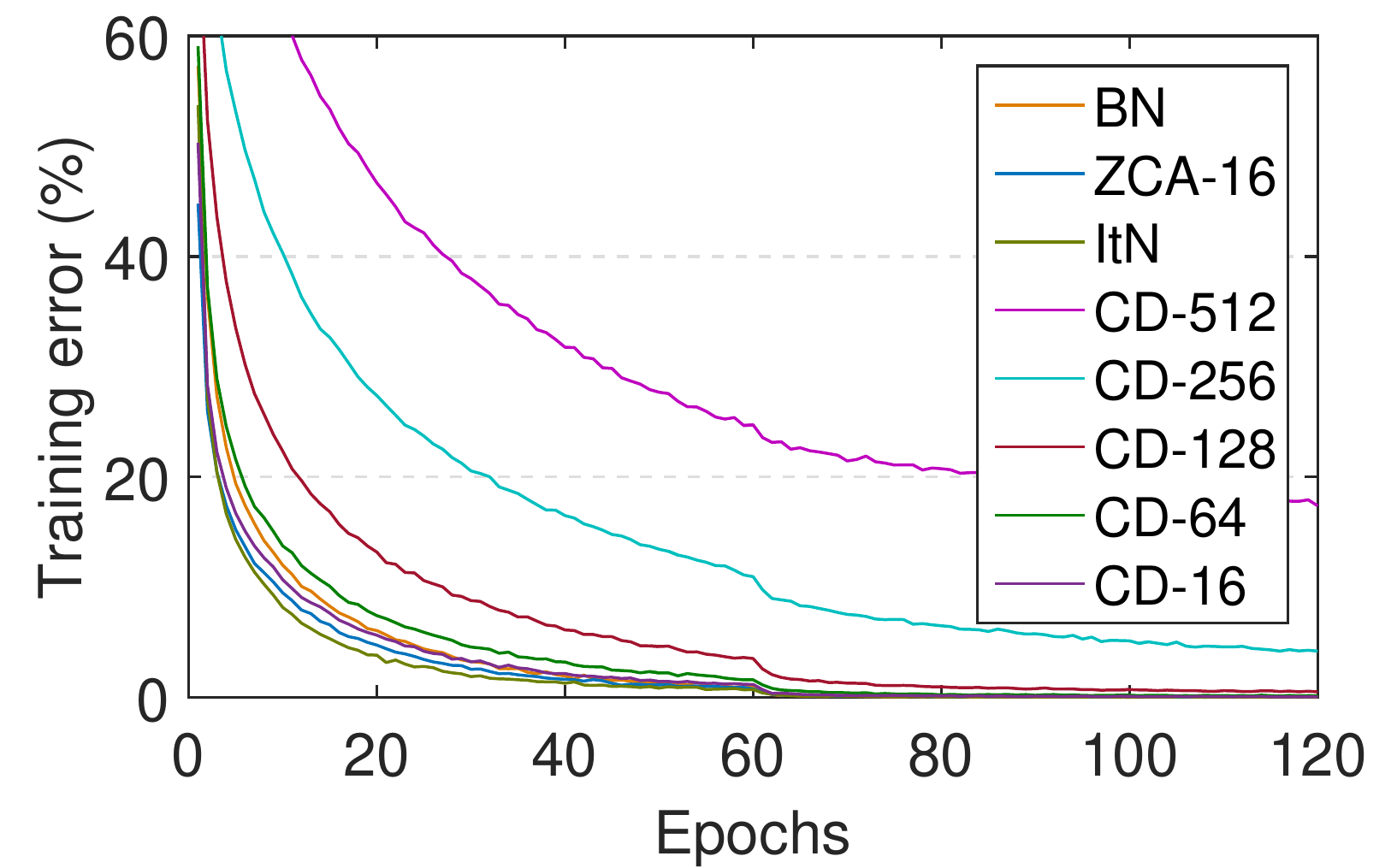}
		\end{minipage}
	}
	\hspace{0.1in}	\subfigure[Test error]{
		\begin{minipage}[c]{.44\linewidth}
			\centering
			\includegraphics[width=4.2cm]{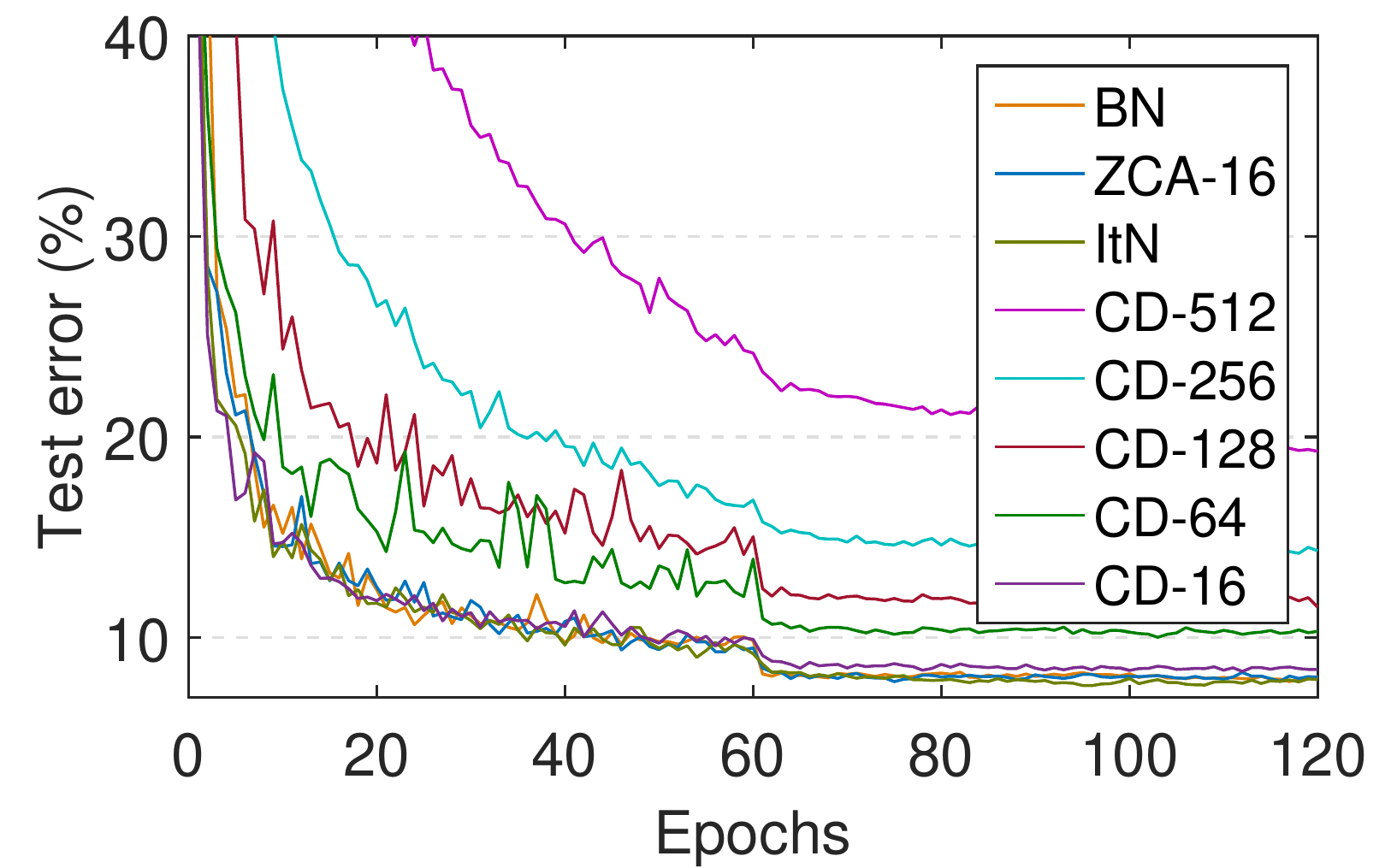}
		\end{minipage}
	}
	\vspace{-0.06in}
	\caption{Experiments on VGG for CIFAR-10 classification. }
	\label{fig:VGG_train}
	\vspace{-0.18in}
\end{figure}
The main experimental observations include: 1) `PCA' fails training under all the configurations, which means that either the training loss does not decrease  or numerical instability appears. This observation is consistent with the previous MLP models. 2) `ZCA-16' can train well, while other `ZCA' related configurations fail training due to numerical instability. This is caused by the back-propagation through the eigen decomposition, which requires different eigenvalues for the mini-batch covariance matrix \cite{2018_CVPR_Huang,2019_ICLR_Siaroin}. 3) `CD' has no numerical instability and can ensure full feature whitening of the models. Fully whitening features by `CD' have significantly worse performance than the group-based ones shown in Figure \ref{fig:VGG_train}. This again suggests that it is essential to control the extent of the whitening for discriminative models. 4) We find that `ItN' (the approximates of the ZCA whitening) works the best.

\begin{figure*}[t]
	\vspace{-0.24in}
	\centering
	\begin{minipage}[c]{.98\linewidth}
		\centering
		\includegraphics[width=14.5cm]{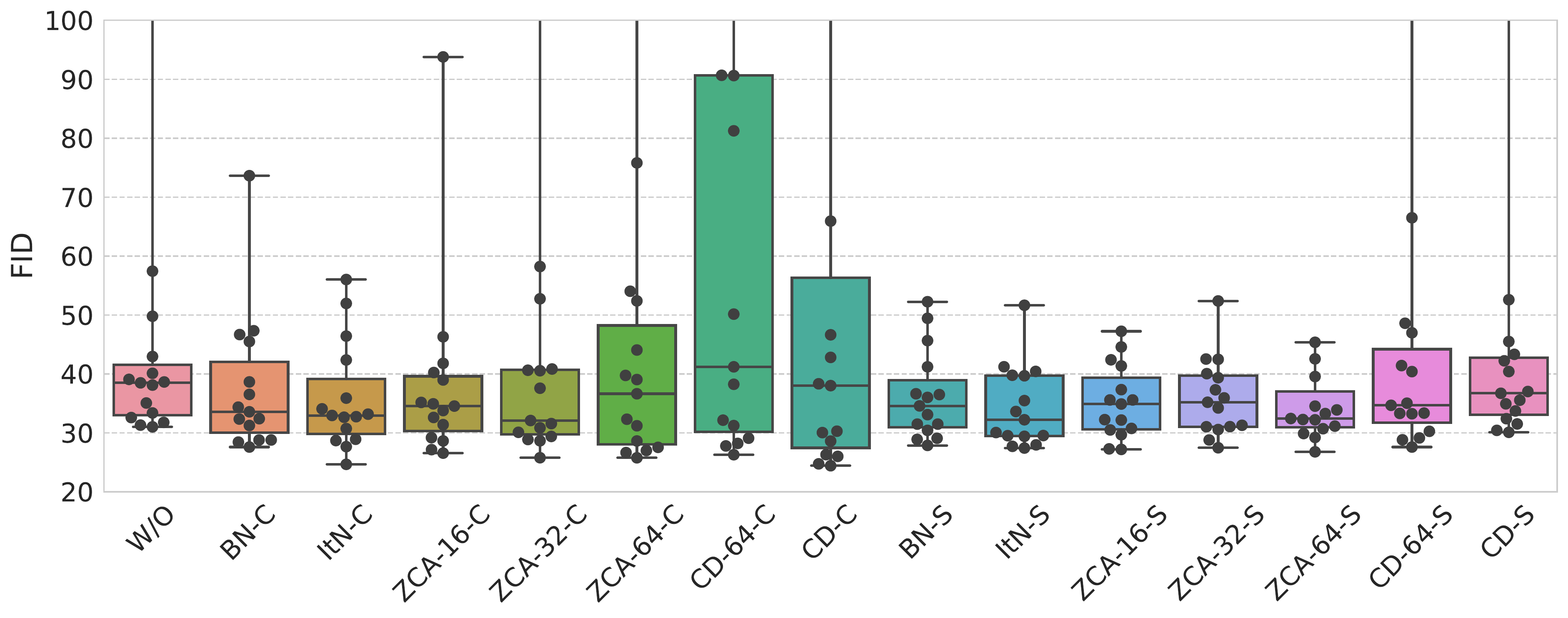}
	\end{minipage}
	\caption{Stability experiments on GAN with hinge loss \cite{2018_ICLR_Miyato,2019_ICLR_Brock} for unconditional image generation. The box shows the quartiles while the whiskers extend to show the rest of the distribution (we limit the FID ranging in  (20,100) for better representation, and we show the full range of FID in the \TODO{Appendix}~\ref{sec:sup-GAN-Stability}).  Note that all methods use covariance matrix to estimate population statistics.}
	\label{fig:GAN_Stability}
	\vspace{-0.22in}
\end{figure*}

	\vspace{-0.15in}
\subsubsection{Residual Network on ImageNet}
	\vspace{-0.05in}
\label{exp_imagenet}

We investigate the effectiveness of all kinds of whitening algorithms on residual networks for ImageNet classification with 1000 classes ~\cite{2009_ImageNet}.
We use the given official 1.28M training images as a training set, and evaluate the top-1 accuracy on the validation
set with 50k images.
\vspace{-0.18in}
\paragraph{Ablation Study on ResNet-18} We first execute an ablation study on the 18-layer residual network (ResNet-18) to explore multiple positions for replacing BN with BW. We consider three architectures:
 1) \textbf{ARC$_A$}:  where we only replace the first BN module of ResNet-18  proposed in \cite{2018_CVPR_Huang};
 2)\textbf{ ARC$_B$}: where we further plug-in the BW layer after the last average pooling (before the last linear layer) to learn the decorrelated feature representations based on \textbf{ARC$_A$}, as proposed in \cite{2019_CVPR_Huang};
 3)\textbf{ ARC$_C$}: where we also replace the $\{2n, n=1,2,...\}$th BN modules (the $\{3n\}$th for ResNet-50) based on \textbf{ARC$_B$}.
 We compare all the whitening transformations and estimation objects in Table \ref{table:Alogorithm}.  `ZCA' and `ZCA$_\Sigma$' denote the ZCA whitening that estimate the population statistics using $\WM{}$ and $\CM{}$, respectively.
 For `PCA', `ZCA' and `CD', we use a group size in $\{16, 64 \}$ and report the best performance from these two configurations.

\begin{table}[t]
	\centering
	\begin{footnotesize}
		\begin{tabular}{c|ccc}
			\bottomrule[1pt]
			Method     & ARC$_A$   & ARC$_B$ & ARC$_C$ \\
			\hline
			Baseline \cite{2015_CVPR_He} &  70.31 & -- & --  \\
			PCA \cite{2018_CVPR_Huang} & 59.93 ($\downarrow$10.38) & -- & -- \\
			PCA$_\Sigma$  & 70.01 ($\downarrow$0.30) & -- & -- \\
			ZCA  \cite{2018_CVPR_Huang}  & 70.58 ($\uparrow$0.27)  & -- & --  \\
			ZCA$_\Sigma$   & 70.62 ($\uparrow$0.31) & -- & --  \\
			CD    &  70.46 ($\uparrow$0.15) & 70.80 ($\uparrow$0.49) & 68.15 ($\downarrow$2.16)   \\
			CD$_\Sigma$ \cite{2019_ICLR_Siaroin}   & 70.55 ($\uparrow$0.24) & 70.89 ($\uparrow$0.58) & 68.56 ($\downarrow$1.75)  \\
			ItN \cite{2019_CVPR_Huang}    & 70.62 ($\uparrow$0.31) & 71.14 ($\uparrow$0.83) &71.26 ($\uparrow$0.95)\\		
			ItN$_\Sigma$     & 70.63 ($\uparrow$0.32) & 71.33 ($\uparrow$1.02) &71.62 ($\uparrow$1.31) \\
			\toprule[1pt]
		\end{tabular}
		\caption{Comparison of validation accuracy ($\%$, single model and
			single-crop) on an 18-layer residual network
			for ImageNet.}
		\label{table:ImageNet-Res-18}
	\end{footnotesize}
		\vspace{-0.22in}
\end{table}

We follow the same experimental setup as described in ~\cite{2015_CVPR_He}, except that we use one GPU and train over 100 epochs. We apply SGD with a mini-batch size of 256, momentum of 0.9 and weight decay of 0.0001. The initial learning rate is set to 0.1 and divided by 10 at 30, 60  and 90 epochs.

The results are shown in Table \ref{table:ImageNet-Res-18}. For \textbf{ARC$_A$}, we find that all whitening methods, except PCA related methods, improve the performance over the baselines. We observe that ZCA (ZCA$_\Sigma$) and its approximate ItN (ItN$_\Sigma$), achieve slightly better performance than CD (CD$_\Sigma$), and this observation is consistent with the results in \cite{2019_ICLR_Siaroin}. This suggests ZCA  whitening, which minimizes the distortion introduced by whitening under the L2 distance \cite{2018_CVPR_Huang}, usually works better than other whitening methods for discriminate classification tasks.
Under \textbf{ARC$_B$} and \textbf{ARC$_C$}, ZCA/PCA related methods suffer from numerical instability. We also observe that CD-related methods have significantly degenerated performance under \textbf{ARC$_C$}. This implies that the  stochasticity introduced by decorrelating multiple layers with CD whitening harms the learning. We find that ItN-related methods can effectively control the stochasticity and achieve further performance improvement on \textbf{ARC$_C$}.  We try a ResNet-18 where all BN layers are replaced with ItN. However, the network has no performance improvement over \textbf{ARC$_A$}, while the  computations introduced are significant, as already observed in \cite{2019_CVPR_Huang}.  These results demonstrate that controlling the extent of whitening (stochasticity) is important for achieving performance improvements over the standardization.

From all the architectures and whitening methods, we observe that using $\Sigma$ to estimate the population statistics is better than using $\WM{}$, especially on \textbf{ARC$_C$}.

\vspace{-0.18in}
\paragraph{Results on ResNet-50} Based on the above observation, we further apply the ItN$_\Sigma$ to ResNet-50. In addition to the standard step learning rate decay \cite{2015_CVPR_He} used in the previous setup, we also consider the cosine learning rate decay \cite{2017_ICLR_Ilya} that is also a basic setting~\cite{2019_CVPR_He} when training on ImageNet, and we want to address that the improvement of the proposed method can be obtained under different setups. For cosine decay, we start with a learning rate of 0.1 and decay it to 0.00001 over 100 epochs. 
We find that the proposed models significantly improves the performance over the original one, under all configurations. The additional time cost of `ItN$_\Sigma$-ARC$_B$' and `ItN$_\Sigma$-ARC$_C$' is $7.03\%$ and $30.04\%$ over the original. 
Note that ItN$_\Sigma$ consistently provides a slight improvement (around $0.1 $ to $0.4$ for all configurations in this experiment) over the original ItN proposed in \cite{2019_CVPR_Huang}, and please see \TODO{Appendix}~\ref{sec:sup-Classification} for details. Moreover, ItN$_\Sigma$ can bear a larger group size and smaller batch size than ItN, which is particularly important in the scenario of training GANs, as we will discuss.

\begin{table}[t]
	\centering
	\begin{small}
		\begin{tabular}{c|c|c}
			\bottomrule[1pt]
			Method   & Step decay   & Cosine decay  \\
			\hline
			Baseline  &  76.20  &  76.62    \\
			ItN$_{\Sigma}$-ARC$_B$  & 77.18 ($\uparrow$0.98) &77.68 ($\uparrow$1.06) \\
			ItN$_{\Sigma}$-ARC$_C$  & \textbf{77.28} ($\uparrow$1.08) &\textbf{77.92} ($\uparrow$1.30)\\
			\toprule[1pt]
		\end{tabular}
		\caption{Results using ResNet-50 for ImageNet. We evaluate the top-1 validation accuracy  ($\%$, single model and	single-crop). }
		\label{tab:res50}
	\end{small}
	\vspace{-0.22in}
\end{table}


	\vspace{-0.05in}
\subsection{Investigation on Training GANs}
	\vspace{-0.05in}
In this section, we provide an empirical analysis on BW algorithms for training GANs. We focus on investigating the effects of different BW algorithms in stabilizing the training and improving  the performance. We evaluate on unconditional image generation for the CIFAR-10 dataset. In our qualitative evaluation of the generated samples, we use the two most common metrics: Fr\'{e}chet Inception Distance (FID) \cite{2017_NIPS_Heusel} (the lower the better) and Inception Score (IS) \cite{2016_NIPS_Salimans} (the higher the better). Following the setup in \cite{2019_ICLR_Siaroin}, we only apply the BW on the generator. Unless otherwise noted, the batch whitening methods used in this section use the covariance matrix $\Sigma$ to estimate the whitening matrix $\HWM{}$.

\begin{table}[t]
	\centering
		\vspace{-0.1in}
	\begin{footnotesize}
		\begin{tabular}{c|cc|cc}
			\bottomrule[1pt]
			& \multicolumn{2}{c|}{Scale $\&$ Shift} &    \multicolumn{2}{c}{Coloring}  \\
			Method   & IS   & FID  & IS   &  FID  \\
			\hline
			BN   & 7.243 $\pm$ 0.073  &  27.89   & 7.317 $\pm$ 0.093 & 27.61  \\
			CD   & 6.986 $\pm$ 0.065  &  30.11   & \textbf{7.612 $\pm$ 0.112} & \textbf{24.44}  \\
			ZCA-64  & \textbf{7.265 $\pm$ 0.069}   &  \textbf{26.8}    & 7.412 $\pm$ 0.118          & 25.79\\
			ItN    &7.246 $\pm$ 0.104        &27.44 & 7.599 $\pm$ 0.089  & 24.68\\
			\toprule[1pt]
		\end{tabular}
		\caption{Results on stability experiments shown in Section \ref{sec:stability}. We report the best FID and the corresponding IS from all the configurations  for different whitening methods.}
		\label{tab:GAN_Stability_performance}
	\end{footnotesize}
	\vspace{-0.1in}
\end{table}

\begin{figure}[]
	\centering
	\begin{minipage}[c]{.98\linewidth}
		\centering
		\includegraphics[width=7.4cm]{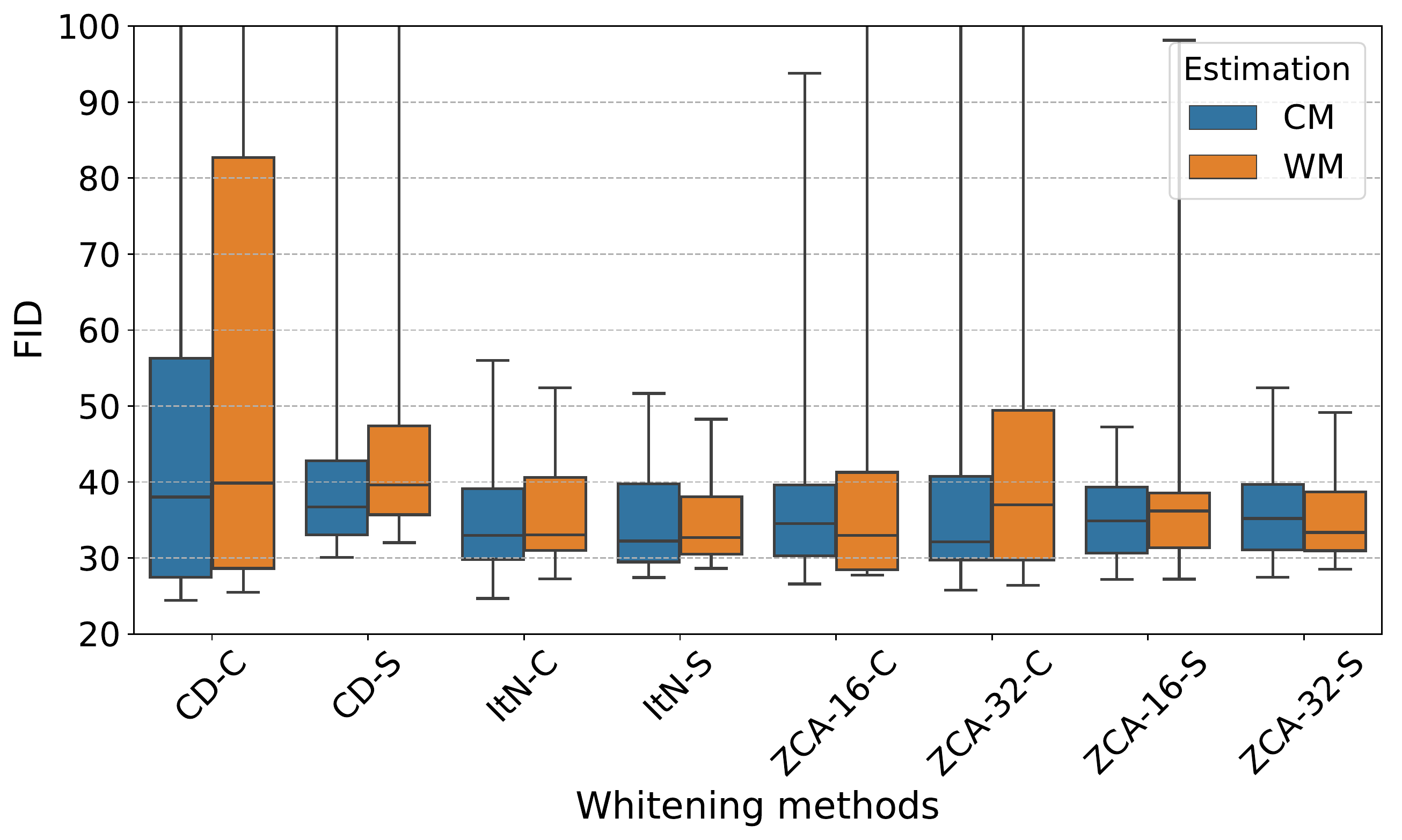}
	\end{minipage}
	\caption{Comparison of BW methods between using Covariance Matrix (CM) and Whitening Matrix (WM) as estimation object.}
	\label{fig:GAN_Estimation}
	\vspace{-0.18in}
\end{figure}

	\vspace{-0.15in}
\subsubsection{Stability Experiments}
	\vspace{-0.05in}
\label{sec:stability}
Following the analysis in \cite{2019_ICML_Kurach}, we conduct experiments to demonstrate the effects of BW algorithms in stabilizing the training of GANs. We use  DCGAN \cite{2016_ICLR_Radford} architectures and use the hinge loss \cite{2018_ICLR_Miyato,2019_ICLR_Brock}.   We provide the implementation details in the \TODO{Appendix}~\ref{sec:sup-GAN}.

 We use the Adam optimizer \cite{2014_CoRR_Kingma} and train for 100 epochs.
We consider 15 hyper-parameter settings (See \TODO{Appendix}~\ref{sec:sup-GAN-Stability} for details) by varying the learning rate $\alpha$, first and second momentum   ($\beta_1$, $\beta_2$) of Adam, and number of discriminator updates per generator update $n_{dis}$.
We use both the `Scale$\&$Shift' (denoted as `-S' ) and `Coloring' (denoted as `-C' ) to recover the representation of BW.
 We calculate the FID distribution of the trained models under 15 configurations, and the results are shown in Figure \ref{fig:GAN_Stability}. The lower the variance, the more stable the model from an optimization perspective.
 In Table \ref{tab:GAN_Stability_performance},
 we also provide the best FID and the corresponding IS from all the configurations  for different whitening methods. Note that full ZCA whitening also suffers from the numerical instability in this experiment, and we thus use the group-based ZCA whitening.

 We observe that `CD' whitening combined with coloring transformation (CD-C) achieves the best performances. However, it has the worst stability over all whitening methods.  We argue that full feature whitening  introduces strong stochasticity, which benefits the diversity of the generated examples. However, the strong stochasticity also harms the training and makes the model sensitive to  the hyperparameter. We find that the coloring operation can improve the performances, especially for the whitening transformations (\eg, BN benefits less). Nevertheless, it also makes the training unstable and sensitive to  the hyperparameter. Besides, we observe that `ZCA-64' achieves better performance than 'CD-64`. This suggests that the advantage of ZCA whitening over CD in discriminative scenarios still exists when training GANS.
  We argue that controlling the magnitude of whitening (stochasticity) is also still important in training GANs. 
   For example, for ItN, the approximation of ZCA whitening has nearly the same performance as CD (Table \ref{tab:GAN_Stability_performance}), and shows better stability (Figure \ref{fig:GAN_Stability}) due to its effective control of the stochasticity, as explained in \cite{2019_CVPR_Huang}.

 We also have obtained similar results when using the non-saturated loss \cite{2014_NIPS_goodfellow}, as shown in the \TODO{Appendix}~\ref{sec:sup-GAN-Stability}.
 
\vspace{-0.16in}
 \paragraph{Comparison of Estimation Object} 
  Considering that the CD whitening in \cite{2019_ICLR_Siaroin} uses the covariance matrix to indirectly estimate the whitening matrix  when training a GAN,  we also compare the whitening methods with different estimation objects. The results are shown in Figure \ref{fig:GAN_Estimation}. We observe that the methods using the covariance matrix as the estimation object,  achieve consistently better FID scores and are generally more stable than using  the whitening matrix. 
 

\begin{table}[t]
	\centering
		\vspace{-0.1in}
	\begin{footnotesize}
		\begin{tabular}{c|cc|cc}
			\bottomrule[1pt]
			& \multicolumn{2}{c|}{DCGAN} &    \multicolumn{2}{c}{ResNet}  \\
			Method   & IS   & FID  & IS   &  FID  \\
			\hline
			CD   & \textbf{7.820 $\pm$ 0.099}  &  22.89   & \textbf{8.522 $\pm$ 0.113} & 18.34  \\
			ZCA-64  &7.672 $\pm$ 0.068 &22.91& 8.492 $\pm$ 0.106 & 17.86\\
			ItN  &7.652 $\pm$ 0.070 &\textbf{22.5}&8.375 $\pm$ 0.093 & \textbf{17.55}\\
			\toprule[1pt]
		\end{tabular}
		\caption{Performance comparison between different whitening methods on the DCGAN and ResNet architectures used in \cite{2019_ICLR_Siaroin}. For fairness, we report all the results at the end of the training.}
		\label{tab:GAN_performance}
	\end{footnotesize}
	\vspace{-0.2in}
\end{table}

	\vspace{-0.15in}
\subsubsection{Validation on Larger Architectures}
\label{sec:larger}
	\vspace{-0.05in}
Based on our previous analysis, we observe that ItN and ZCA whitening (with an appropriate group size)  have similar performance to CD in training GANs, when using the covariance matrix as the estimation object. We further apply the ZCA whitening and ItN to the DCGAN and Resnet models, where CD whitening achieves nearly the state-of-the-art performance on CIFAR-10. 
We use the code provided in \cite{2019_ICLR_Siaroin} and use the same setup as in \cite{2019_ICLR_Siaroin}. We replace all the CD whitening in the models with ItN/ZCA whitening. 
The results are shown in Table \ref{tab:GAN_performance}. 
We observe that the CD whitening has no significant advantages over ItN/ZCA-G64. Ultimately, ItN achieves the best performances in terms of the FID evaluation, while CD whitening is the best in terms of IS.   This suggests that there is still room to improve the performance of batch whitening in training GANs by further finely controlling the stochasticity.  

\vspace{-0.08in}
\section{Conclusions}
	\vspace{-0.05in}
In this paper, we provided a stochasticity analysis of batch whitening that thoroughly explains why different whitening transformations show significant differences in performance when training DNNs, despite their equivalent improvement of the conditioning.  Our analysis provides insights for designing new normalization algorithms and constructing new network architectures. We believe that our analysis will open new avenues of research in better understanding normalization over batch data.

\vspace{0.1in}
\noindent\textbf{Acknowledgement}
We  thank Anna Hennig and Ying Hu for their help with proofreading.

{\small
	\bibliographystyle{ieee_fullname}
	\bibliography{3whitening}
}

\clearpage

\appendix
\renewcommand{\thealgorithm}{\Roman{algorithm}}
\setcounter{algorithm}{0}

\renewcommand{\thetable}{A\arabic{table}}
\setcounter{table}{0}

\renewcommand{\thefigure}{A\arabic{figure}}
\setcounter{figure}{0}

\section{Back-propagation}
\label{sec-sup-BP}
Given the Batch Whitening (BW) algorithms described in Algorithm \ref{alg_forward} of Section~\ref{sec:experimental-landscape}, we provide the corresponding back-propagation during training in Algorithm \ref{alg_backward}. Note that it is easy to calculate $\D{\NX{}} $ in Line.~3 of Algorithm \ref{alg_backward}, given the specific recovery operation shown in the  paper. Here, we mainly elaborate on the nontrivial solution of back-propagation through the whitening transformation, \ie, how to calculate $\D{\CM{}}$ given $\D{\WM{}}$, shown in Line. 5 of  Algorithm \ref{alg_backward}.

We consider the following whitening transformations discussed in the paper: Principal Component
Analysis (PCA) whitening \cite{2018_CVPR_Huang}, Zero-phase Component
Analysis (ZCA) whitening \cite{2018_CVPR_Huang}, Cholesky Decomposition (CD) whitening \cite{2019_ICLR_Siaroin} and ItN \cite{2019_CVPR_Huang} that approximates ZCA whitening. The back-propagations of all whitening transformations are derived in their original papers \cite{2018_CVPR_Huang, 2019_ICLR_Siaroin, 2019_CVPR_Huang}. Here, we provide the results for completeness.

\begin{algorithm}[tb]
	\algsetup{linenosize=\footnotesize}
	\footnotesize
	\caption{Back-propagation of batch whitening algorithms.}
	\label{alg_backward}
	\begin{algorithmic}[1]
		\begin{small}
			\STATE \textbf{Input}: mini-batch gradients $ \D{\Y{}} \in \mathbb{R}^{d \times m} $.
			\STATE \textbf{Output}: $ \D{\X{}} \in \mathbb{R}^{d \times m}$.
			\STATE  Calculate: $\D{\NX{}} = \D{\Y{}} \frac{\partial \phi_3(\NX{})}{\partial \NX{}}$.
			\STATE  Calculate: $\D{\WM{}}= \D{\NX{}} \X{}^T$.
			\STATE  Calculate: $\D{\CM{}}= \D{\WM{}} \frac{\partial \phi_1(\CM{})}{\partial \CM{}} $.
			\STATE  Calculate: $\D{\X{}}=\WM{}^T \D{\NX{}}+ \frac{1}{m}(\D{\CM{}} +\D{\CM{}}^T) \X{} $.

			%
		\end{small}
	\end{algorithmic}
\end{algorithm}

\subsection{PCA Whitening}
PCA Whitening uses $\WM{PCA}=\Lambda^{-\frac{1}{2}} \mathbf{D}^T$, where  $\Lambda=\mbox{diag}(\sigma_1, \ldots,\sigma_d)$ and $\mathbf{D}=[\mathbf{d}_1, ...,
\mathbf{d}_d]$ are the eigenvalues and associated eigenvectors of $\Sigma$, \ie $\Sigma = \mathbf{D}
\Lambda \mathbf{D}^T$. 


The backward pass of PCA whitening is:
\begin{align}
\label{eq_dLdLambda}
\D{\Lambda}&=(\D{\WM{PCA}}) \mathbf{D} (-\frac{1}{2} \Lambda^{-3/2} )\\
\label{eq_dLdD}
\D{ \mathbf{D}}&=  ( \D{\WM{PCA}})^T \Lambda^{-1/2}  \\
\label{eq_dLdSigma}
\D{\Sigma}&=\mathbf{D}\{ (\mathbf{K}^T \odot (\mathbf{D}^T \D{ \mathbf{D}} )) + (\D{\Lambda})_{diag}\} \mathbf{D}^T,
\end{align}
\hspace{-0.05in}where  $\mathbf{K} \in \mathbb{R}^{d\times d}$ is 0-diagonal with
$\mathbf{K}_{ij}= \frac{1}{\sigma_i  -\sigma_j} [i\neq j]$,
the $\odot$ operator is an element-wise matrix multiplication, and $(\D{\Lambda})_{diag}$ sets the off-diagonal elements of $\D{\Lambda}$ to zero.

\subsection{ZCA Whitening}
ZCA whitening uses $\WM{ZCA}=\mathbf{D} \Lambda^{-\frac{1}{2}} \mathbf{D}^T$, where the  PCA whitened input is rotated back by the corresponding rotation matrix $\mathbf{D}$.

The backward pass of ZCA whitening is:
\begin{align}
\D{\Lambda}&=\mathbf{D}^T (\D{\WM{ZCA}}) \mathbf{D} (-\frac{1}{2} \Lambda^{-3/2} )\\
\D{ \mathbf{D}}&=  (\D{\WM{ZCA}}+  (\D{\WM{ZCA}})^T)  \mathbf{D} \Lambda^{-1/2} \\
\D{\Sigma}&=\mathbf{D}\{ (\mathbf{K}^T \odot (\mathbf{D}^T \D{ \mathbf{D}} )) + (\D{\Lambda})_{diag}\} \mathbf{D}^T.
\end{align}

\begin{algorithm}[tb]
	\caption{Whitening activations with Newton's iteration.}
	\label{alg_ItN}
	\begin{algorithmic}[1]
		\begin{small}
			\STATE \textbf{Input}: $\CM{}$.
			\STATE \textbf{Output}: $\WM{ItN}$.
			\STATE	$\Sigma_{N}=\frac{\CM{}}{tr(\CM{})}$ .
			\STATE $\mathbf{P}_0=\mathbf{I}$.
			\FOR {$k = 1 ~~to ~~T $}
			\STATE $\mathbf{P}_{k}=\frac{1}{2} (3 \mathbf{P}_{k-1} - \mathbf{P}_{k-1}^{3} \Sigma_{N})$
			\ENDFOR	
			\STATE   $\WM{ItN} = \mathbf{P}_T / \sqrt{tr(\Sigma)}$
		\end{small}
	\end{algorithmic}
\end{algorithm}

\begin{figure}[]
	\centering
	\hspace{-0.15in}	\subfigure[ZCA Whitening]{
		\begin{minipage}[c]{.44\linewidth}
			\centering
			\includegraphics[width=4.2cm]{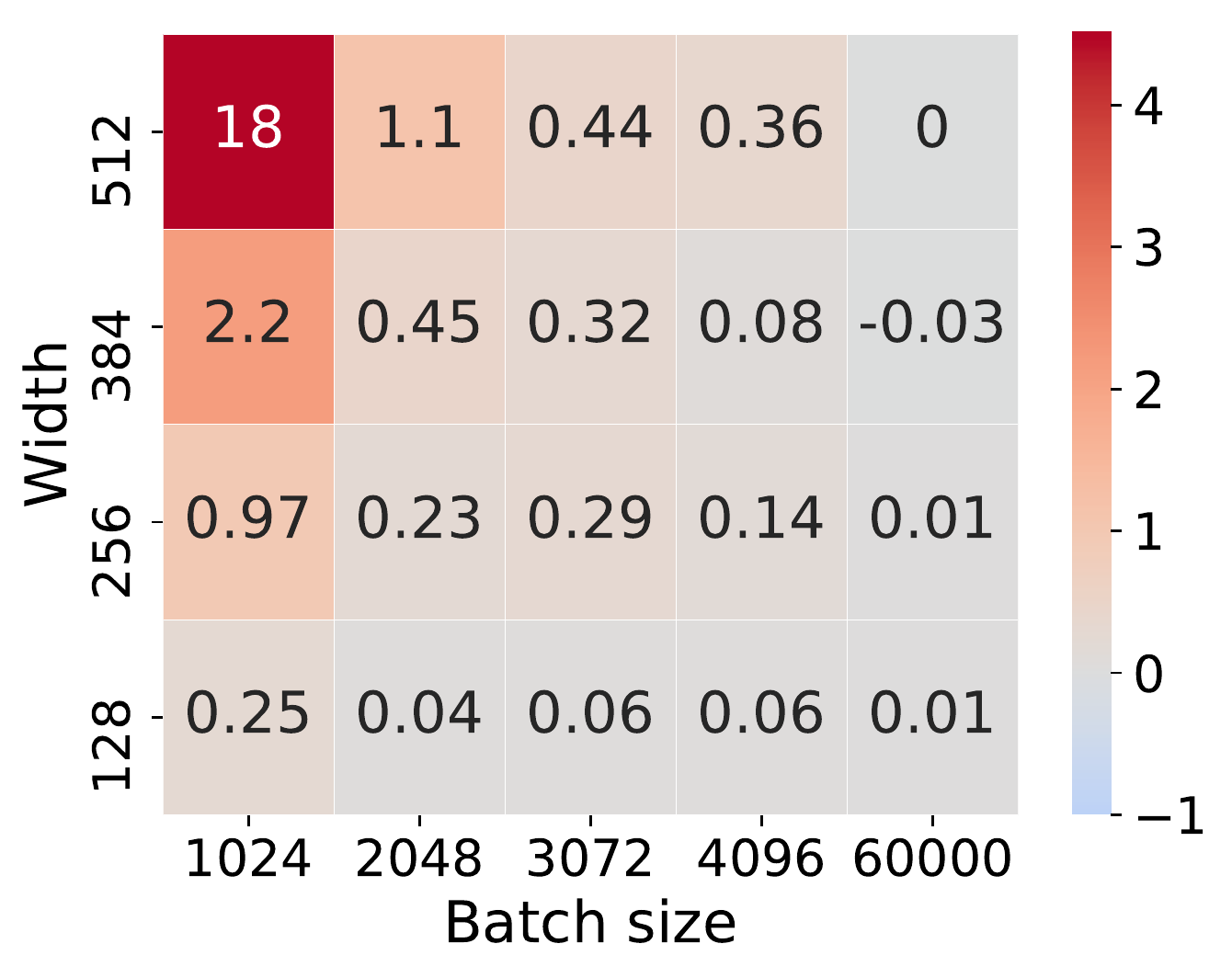}
		\end{minipage}
	}
	\hspace{0.15in}	\subfigure[CD Whitening]{
		\begin{minipage}[c]{.44\linewidth}
			\centering
			\includegraphics[width=4.2cm]{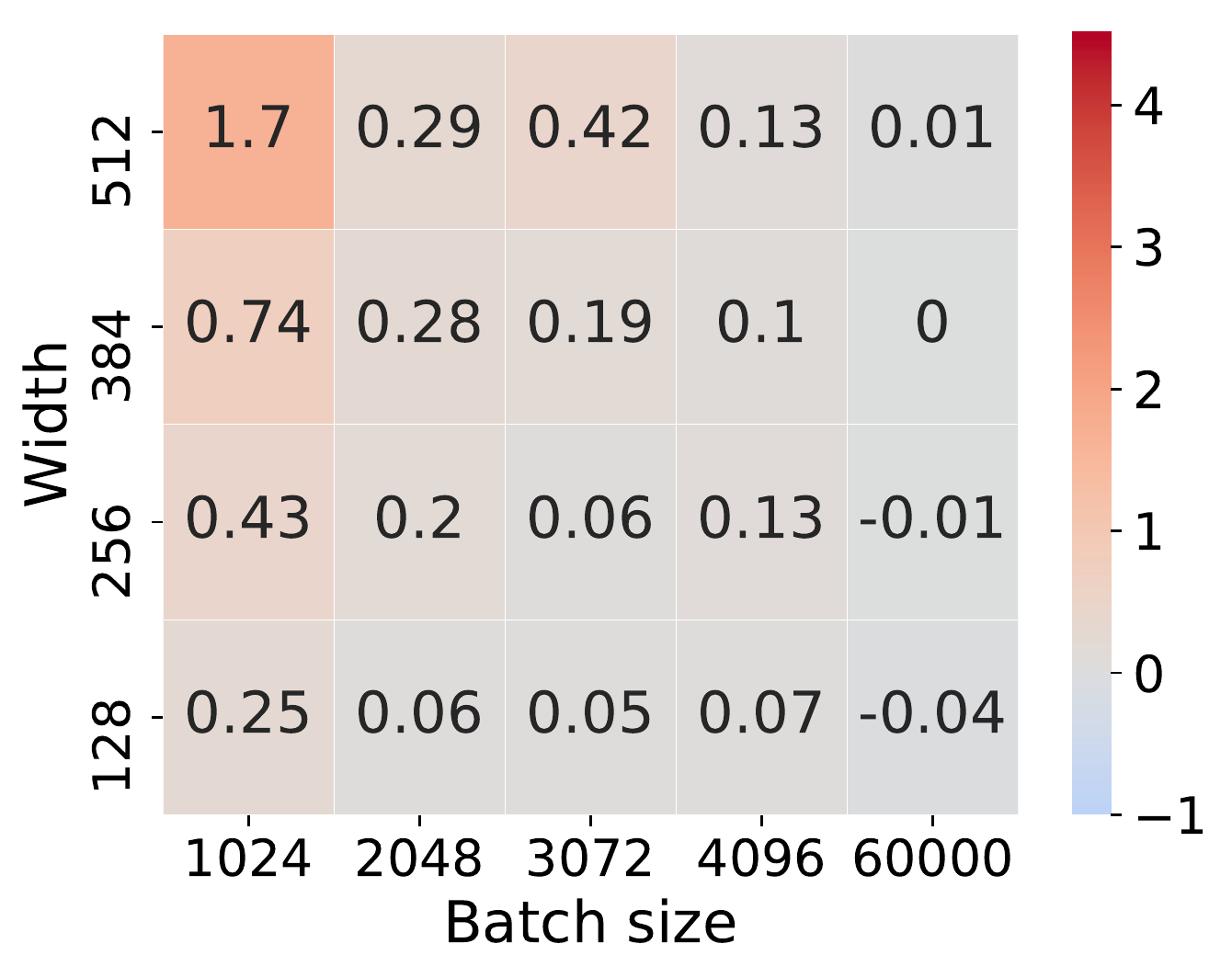}
		\end{minipage}
	}
	\caption{Comparison of estimating the population statistics of whiten matrix $\HWM{}$ using different estimation objects ($\WM{}$/$\CM{}$). We train MLPs  with varying widths (the number of neurons in each layer) and batch sizes, for MNIST classification.
		We evaluate the difference in test accuracy between  $\HWM{\CM{}}$ and $\HWM{\WM{}}$: $AC(\HWM{\CM{}})- AC(\HWM{\WM{}})$. (a) and (b) show the results using ZCA and CD whitening respectively, under a learning rate of 0.5. }
	\label{fig:sup-Estimation}
\end{figure}

\subsection{CD Whitening}
CD whitening uses $\WM{CD}=\mathbf{L}^{-1}$, where $\mathbf{L}$ is a lower triangular matrix from the Cholesky decomposition, with $\mathbf{L} \mathbf{L}^T=\CM{}$. 
The backward pass of CD whitening is:

\begin{align}
\D{\mathbf{L}}&= -  \WM{CD}^T \D{\WM{CD}} \WM{CD}^T \\
\D{\CM{}}&=\frac{1}{2}  \mathbf{L}^{-T}(P \odot \mathbf{L}^{T} \D{\mathbf{L}} +(P \odot \mathbf{L}^{T} \D{\mathbf{L}})^T ) \mathbf{L}^{-1}. 
\end{align}
\subsection{ItN Whitening}
ItN, which approximates the ZCA whitening using Newton's iteration, is shown in Algorithm~\ref{alg_ItN}, where $T$ is the iteration number and $tr(\cdot)$ denotes the trace of a matrix.

Given $\D{\WM{ItN}}$, the back-propagation is: 
\begin{small}
	\begin{align}
	\label{eqn:backward-1}
	\D{\mathbf{P}_{T}}=& \frac{1}{\sqrt{tr(\Sigma)}} \D{\WM{ItN}} \nonumber \\
	\D{\Sigma_{N}}=& -\frac{1}{2} \sum_{k=1}^{T} (\mathbf{P}_{k-1}^3)^T \D{\mathbf{P}_k}  \nonumber  \\
	\D{\Sigma} =&  \frac{1}{tr(\Sigma)} \D{\Sigma_{N}}
	-\frac{1}{(tr(\Sigma))^2} tr(\D{\Sigma_{N}}^T \Sigma)   \mathbf{I}   \nonumber \\
	& ~~~-\frac{1}{2(tr(\Sigma))^{3/2}} tr((\D{\WM{ItN}})^T \mathbf{P}_T) \mathbf{I}.
	\end{align}
\end{small}
\hspace{-0.05in}Here, $\D{\mathbf{P}_k}$ can be calculated by the following iterations:
\begin{small}
	
	\begin{align}
	\label{eqn:backward-iteration}
	\D{\mathbf{P}_{k-1}}& =\frac{3}{2} \frac{\partial{L}}{\partial{\mathbf{P}_{k}}}
	-\frac{1}{2} \D{\mathbf{P}_k}  (\mathbf{P}_{k-1}^2 \Sigma_{N})^T
	-\frac{1}{2}  (\mathbf{P}_{k-1}^2)^T  \D{\mathbf{P}_k} \Sigma_{N}^T
	\nonumber \\
	&-  \frac{1}{2}(\mathbf{P}_{k-1})^T \D{\mathbf{P}_k} (\mathbf{P}_{k-1} \Sigma_{N})^T
	,  ~~k=T,...,1.
	\end{align}
\end{small}

\section{Stochasticity During Inference}
\label{sec-sup:trainingAndInfer}
In Section~\ref{sec:sto-infer} of the  paper, we mentioned that we try other learning rates in the experiments to compare the estimation methods during inference. Here, we provide the results under a learning rate of 0.5, shown in Figure~\ref{fig:sup-Estimation}.



\section{More Details for Experiments}

%
%
\subsection{Classification on ImageNet}
\label{sec:sup-Classification}
In Section~\ref{exp_imagenet} of the  paper, we mentioned that ItN$_\Sigma$ consistently provides a slight improvement over the original ItN proposed in \cite{2019_CVPR_Huang} (using $\WM{}$ as an estimation object). Here, we provide the results, including the original ItN, in Table~\ref{tab:sup-res50}. We use an iteration number of 5 for all ItN-related methods. 
For ItN, we use a group size of 64, as recommended in \cite{2019_CVPR_Huang}. 
For ItN$_{\Sigma}$, we use a larger group size of 256, since ItN$_{\Sigma}$ can still provide a good estimation of population statistics in high dimensions during inference, as shown in the  paper. 
\begin{table}[t]
	\centering
	\begin{small}
		\begin{tabular}{c|c|c}
			\bottomrule[1pt]
			Method   & Step decay   & Cosine decay  \\
			\hline
			Baseline  &  76.20  &  76.62    \\
			ItN-ARC$_B$ \cite{2019_CVPR_Huang} & 77.07 ($\uparrow$0.87) &77.47 ($\uparrow$0.85) \\
			ItN-ARC$_C$ \cite{2019_CVPR_Huang}  & 76.97 ($\uparrow$0.77) &77.43 ($\uparrow$0.81)\\
			ItN$_{\Sigma}$-ARC$_B$  & 77.18 ($\uparrow$0.98) &77.68 ($\uparrow$1.06) \\
			ItN$_{\Sigma}$-ARC$_C$  & \textbf{77.28} ($\uparrow$1.08) &\textbf{77.92} ($\uparrow$1.30)\\
			\toprule[1pt]
		\end{tabular}
		\caption{Results using ResNet-50 for ImageNet. We evaluate the top-1 validation accuracy  ($\%$, single model and	single-crop). }
		\label{tab:sup-res50}
	\end{small}
\end{table}

\begin{table}[t]
	\centering
	\begin{footnotesize}
		\begin{tabular}{ll}
			\toprule
			PARAMETER    & VALUE \\
			\hline
			Learning Rate $\alpha$ & $\{0.0001, 0.0002, 0.001\}$ \\
			$(\beta_1,\beta_2, n_{dis} )$ & $\{(0, 0.9, 1), (0.5,0.9,5), (0.5,0.999,1),$\\
			&$(0.5,0.999,5),(0.9,0.999,5)\}$  \\
			\bottomrule
		\end{tabular}
		\vspace{0.1in}
		\caption{The scope of hyperparameter ranges used in the stability experiments. The Cartesian product of the values suffices to uncover most of the recent results from the literature \cite{2019_ICML_Kurach}.  }
		\label{table:sup-StabilityGAN}
	\end{footnotesize}
\end{table}

\begin{table}[t]
	\centering
	\begin{footnotesize}
		\begin{tabular}{c|cc|cc}
			\bottomrule[1pt]
			& \multicolumn{2}{c|}{Scale $\&$ Shift} &    \multicolumn{2}{c}{Coloring}  \\
			Method   & IS   & FID  & IS   &  FID  \\
			\hline
			BN   & 7.141 $\pm$ 0.066  &  30.26   & 7.264 $\pm$ 0.122 & 28.35  \\
			CD   & 7.135 $\pm$ 0.066  &  30.40   & 7.362 $\pm$ 0.069 & \textbf{25.97}  \\
			ZCA-64  & 7.128 $\pm$ 0.067   &  29.24    & \textbf{7.445 $\pm$ 0.094}          & 26.78\\
			ItN    &\textbf{7.369 $\pm$ 0.104}        &\textbf{29.02} & 7.396 $\pm$ 0.054  & 28.24\\
			\toprule[1pt]
		\end{tabular}
		\caption{Results on stability experiments with non-saturating loss. We report the best FID and the corresponding IS from all the configurations  for different whitening methods.}
		\label{tab:sup-GAN_Stability_performance}
	\end{footnotesize}
\end{table}

\begin{figure*}[t]
	\centering
	\begin{minipage}[c]{.98\linewidth}
		\centering
		\includegraphics[width=15.0cm]{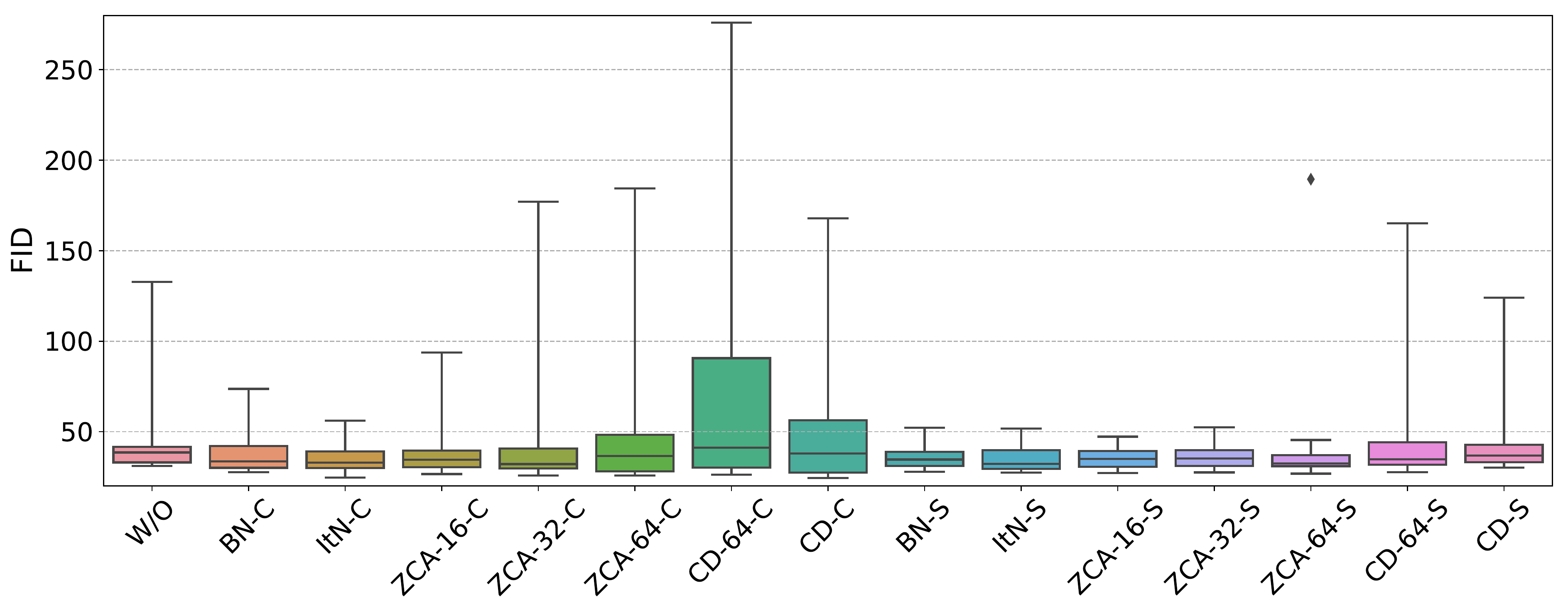}
	\end{minipage}
	\caption{Stability experiments on GAN with hinge loss \cite{2018_ICLR_Miyato,2019_ICLR_Brock} for unconditional image generation. We show the FIDs with full range.}
	\label{fig:sup-GAN_Stability}
\end{figure*}

\begin{figure}[]
	\centering
	\begin{minipage}[c]{.98\linewidth}
		\centering
		\includegraphics[width=7.4cm]{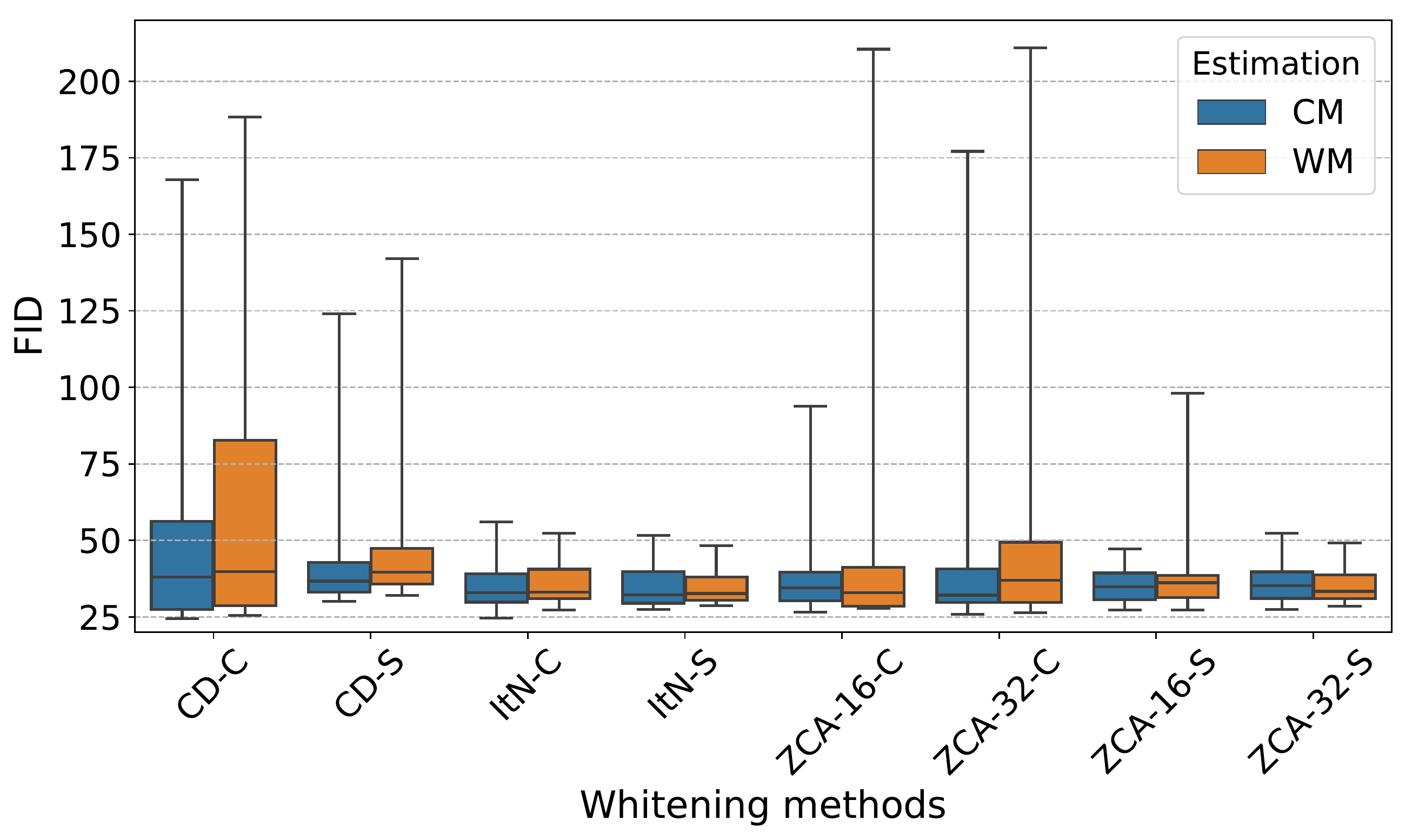}
	\end{minipage}
	\caption{Comparison of BW methods when using the Covariance Matrix (CM) and Whitening Matrix (WM) as the estimation object. We show the FIDs with full range.}
	\label{fig:sup-GAN_Estimation}
\end{figure}

\subsection{Training GANs}
\label{sec:sup-GAN}

\subsubsection{Stability Experiments}
\label{sec:sup-GAN-Stability}
\paragraph{Experimental Setup}
The DCGAN architecture in these experiments is shown in Table \ref{tab:stable dc}. Note that spectral normalization \cite{2018_ICLR_Miyato} is applied on the discriminator, following the setup shown in \cite{2019_ICLR_Siaroin,2018_ICLR_Miyato}.
We evaluate the FID between the 10k test data and 10K randomly generated examples. 
We use the Adam optimizer \cite{2014_CoRR_Kingma} and train for 100 epochs.
We consider 15 hyper-parameter settings (Table \ref{table:sup-StabilityGAN}) by varying the learning rate $\alpha$, first and second momentum   ($\beta_1$, $\beta_2$) of Adam, and number of discriminator updates per generator update $n_{dis}$.
The learning rate is linearly decreased until it reaches 0. 

\paragraph{Results}
In Figures~\ref{fig:GAN_Stability} and~\ref{fig:GAN_Estimation} of the paper, we only show the results with FID ranging in (20, 100) for better representation. Here, we provide the corresponding FIDs with full range in Figure \ref{fig:sup-GAN_Stability} and \ref{fig:sup-GAN_Estimation}. 

\begin{figure*}[]
	\centering
	\hspace{-0.15in}	\subfigure[FID ranging in (20,100)]{
		\begin{minipage}[c]{.88\linewidth}
			\centering
			\includegraphics[width=15.4cm]{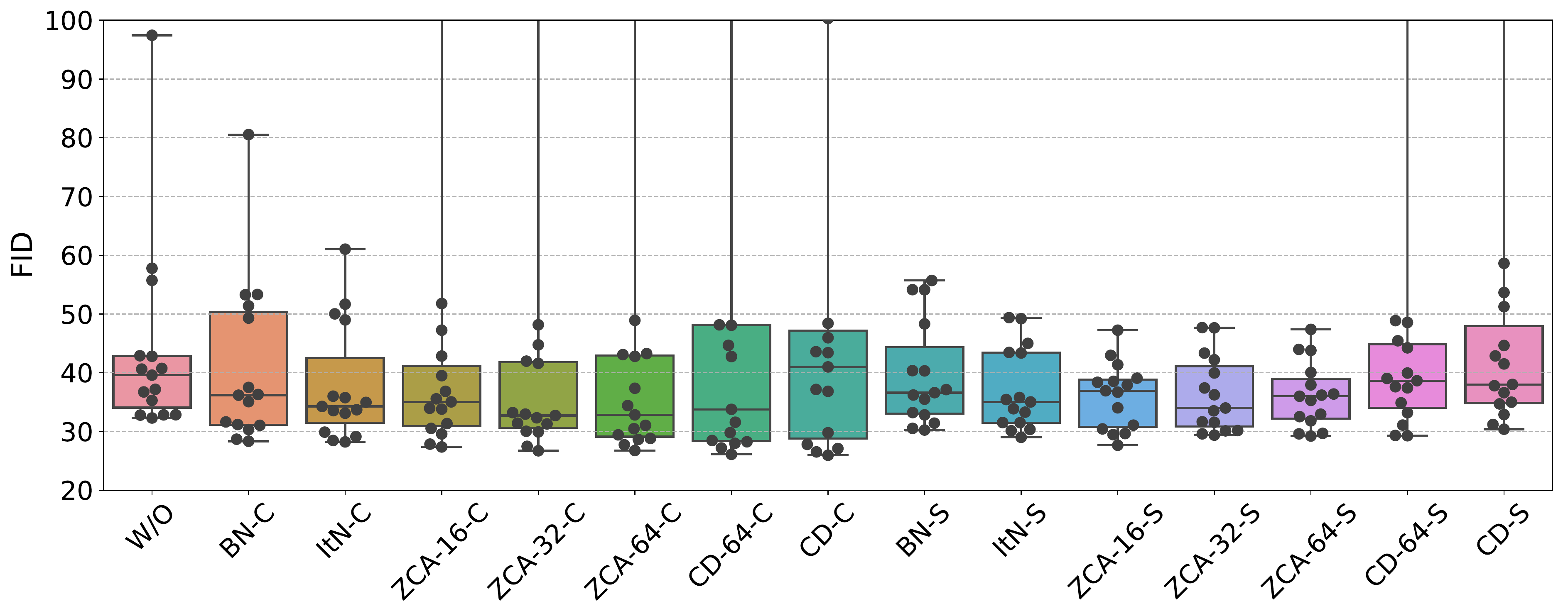}
		\end{minipage}
	}\\
	
	\caption{ Stability experiments on GAN with non-saturating loss \cite{2014_NIPS_goodfellow} for unconditional image generation.}
	\label{fig:sup-GAN_Stability_ns}
\end{figure*}

We perform experiments on DCGAN with the non-saturating loss \cite{2014_NIPS_goodfellow}. Figure \ref{fig:sup-GAN_Stability_ns} shows the results. Here, we obtain similar observations as for the DCGAN with the hinge loss, shown in Section~\ref{sec:stability} of the paper.

\subsubsection{Validation on Larger Architecture}
\label{sec:sup-GAN-Larger}
\paragraph{Experimental Setup}
The architectures in these experiments are shown in Table \ref{tab:large dc} (DCGAN) and \ref{tab:large res} (ResNet). The experimental setup is as follows \cite{2019_ICLR_Siaroin}:
We evaluate the FID between the 10k test data and 10K randomly generated examples. 
We use the Adam optimizer \cite{2014_CoRR_Kingma} with a  learning rate $\alpha=2e^{-4}$. The learning rate is linearly decreased until it reaches 0.  We use  first momentum $\beta_1=0$
and second momentum $\beta_2=0.9$. 
For DCGAN, we use the number of discriminator updates per generator update $n_{dis}=1$ and train the model for 100 epochs, while for ResNet, we use $n_{dis}=5$ and train for 50 epochs. 


\subsection{Comparison in Running Time}
\label{sec-sup:time}
The main disadvantage of whitening methods, compared to  standardization, is their expensive computational costs. 
Here, we provide the time costs of the trained models discussed in the paper. 

We note that the computational cost of the Singular Value Decomposition (SVD) used in ZCA/PCA whitening and the Cholesky Decomposition highly depends on the specific deep learning platform (\eg, PyTorch \cite{2017_NIPS_pyTorch}  or Tensorflow \cite{2016_Tensorflow}). We thus conduct experiments both on PyTorch (version number: 1.0.1) and Tensorflow (version number: 1.13), with CUDA (version number: 10.0.130). 
All implementations of BW are based on the API provided by PyTorch and Tensorflow. 
We hope that our experiments can provide  good guidelines for applying BW in practice.

\subsubsection{Classification Models Using PyTorch}
\begin{figure}[]
	\centering
	\hspace{-0.15in}	\subfigure[ARC$_B$]{
		\begin{minipage}[c]{.44\linewidth}
			\centering
			\includegraphics[width=4.2cm]{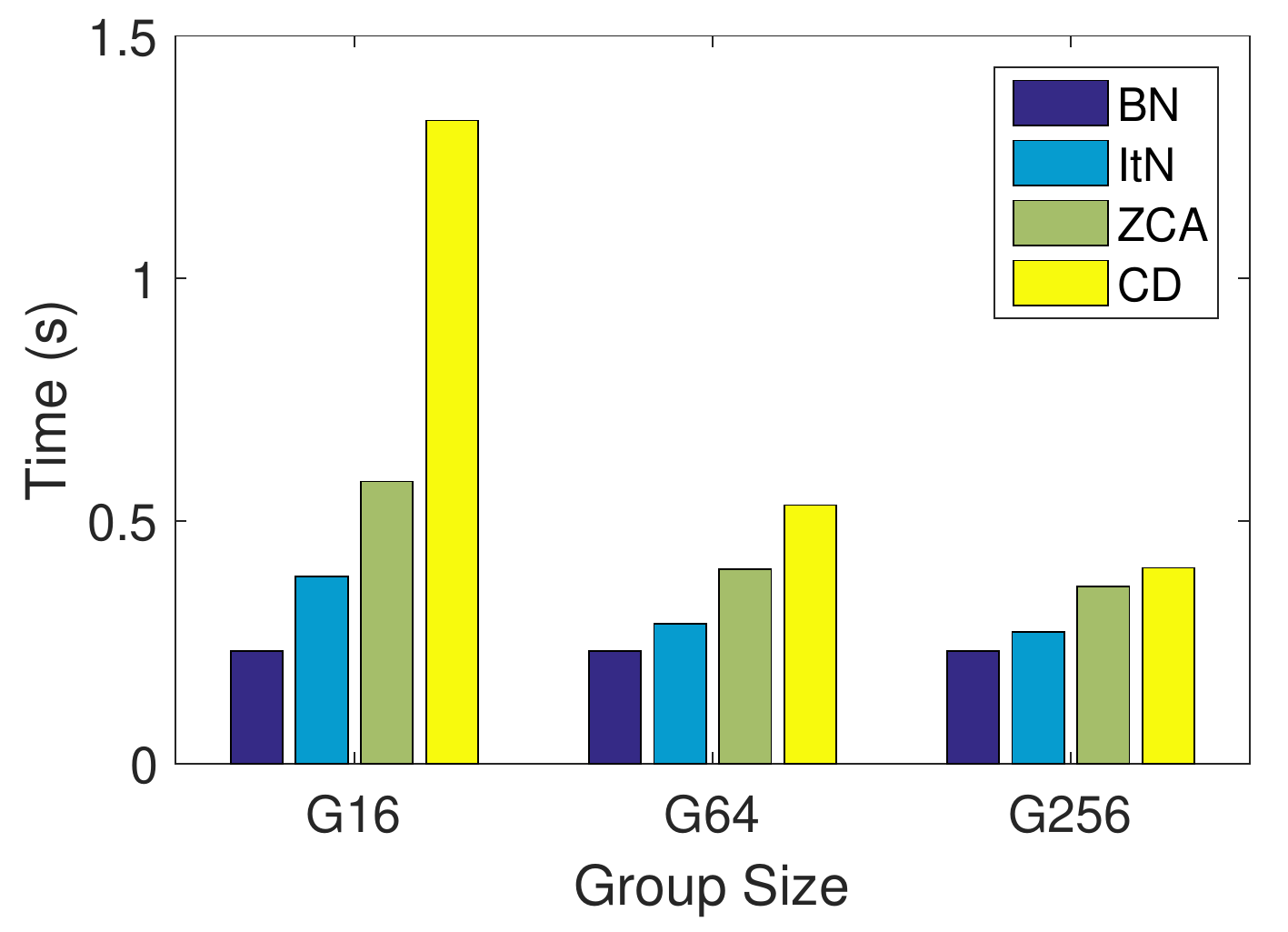}
		\end{minipage}
	}
	\hspace{0.15in}	\subfigure[ARC$_C$]{
		\begin{minipage}[c]{.44\linewidth}
			\centering
			\includegraphics[width=4.2cm]{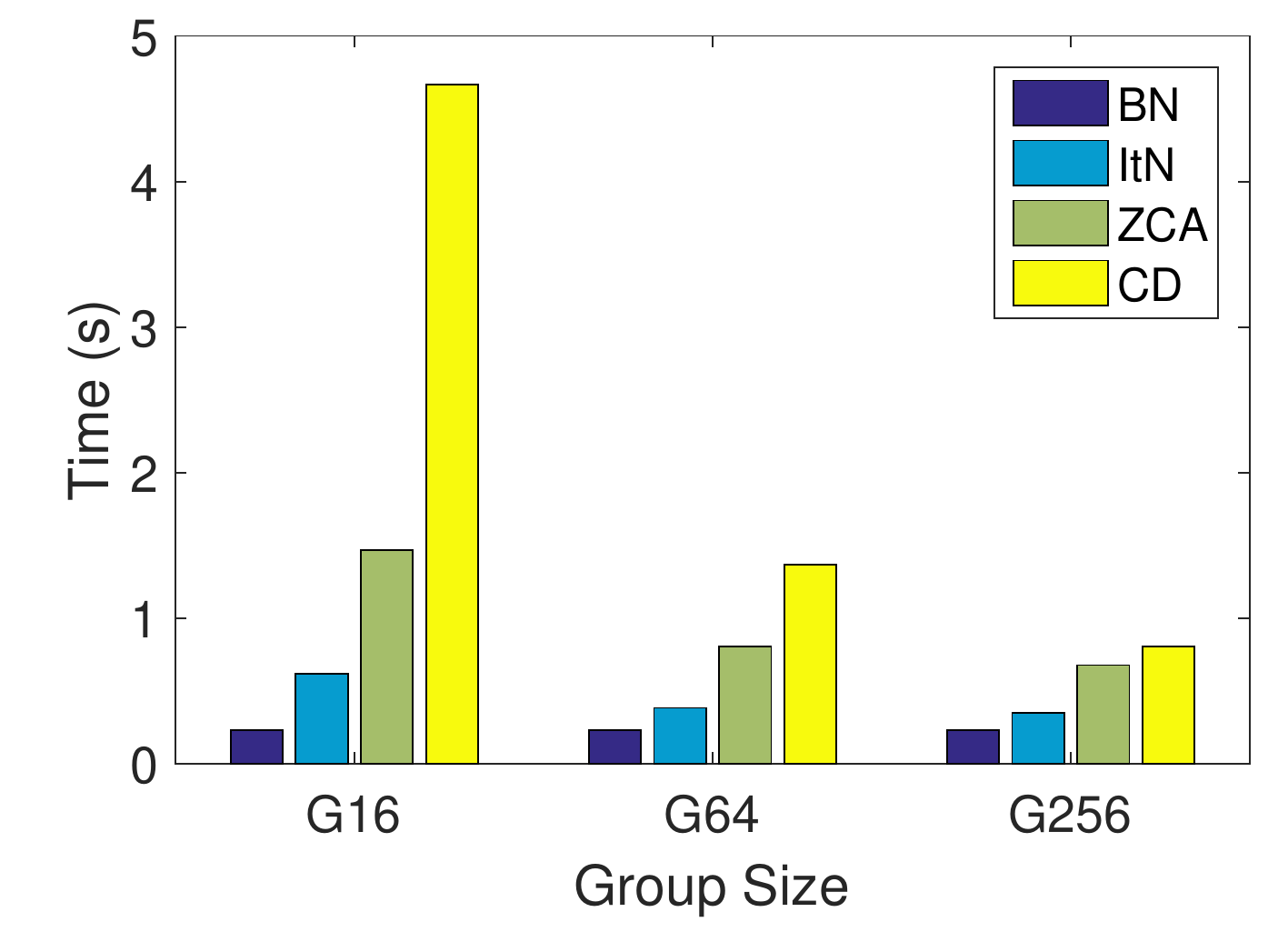}
		\end{minipage}
	}
	\caption{Time cost (s/iteration) on ResNet-18 for ImageNet  classification described in Section~\ref{exp_imagenet} of the paper. We compare the BW methods with different group sizes under (a) ARC$_{B}$ and (b) ARC$_{C}$, and the results are averaged over 100 iterations.}
	\label{fig:Time-res18}
\end{figure}

Table \ref{tab:time vgg} shows the time costs of the VGG models for CIFAR-10 classification, described in Section~\ref{sec:VGG} of the  paper. Figure \ref{fig:Time-res18} and Table \ref{tab:time resnet 50} show the time costs of the ResNet-18 and ResNet-50 described in Section~\ref{exp_imagenet} of the paper, respectively. Note that we run ResNet-18 on one GPU, while ResNet-50 on two GPUs. 

Our main observations include: (1) ItN  is more computationally efficient. (2) CD whitening is slower than ZCA whitening, which is contrary to the observation in \cite{2019_CVPR_Huang} where CD whitening was significantly more efficient than ZCA. The main reason for this is that CD whitening requires the Cholesky decomposition and matrix inverse, which are slower than SVD under the PyTorch platform. 
We note that under Tensorflow, which was used to implement CD whitening in \cite{2019_ICLR_Siaroin}, CD whitening is more efficient than ZCA, as shown in Section \ref{sec:Gan_Tensor}. 
(3) Group based whitening with small groups requires more training time than larger groups. The main reason is our unoptimized implementation where we whiten each group sequentially instead of in parallel, because the corresponding  platform does not provide an easy way to use linear algebra library of CUDA in parallel, which is also mentioned in \cite{2018_CVPR_Huang}. We believe BW would be more widely used if the group based methods could be easily implemented in parallel.  


\begin{figure}[]
	\centering
	\hspace{-0.15in}	\subfigure[GPU]{
		\begin{minipage}[c]{.44\linewidth}
			\centering
			\includegraphics[width=4.2cm]{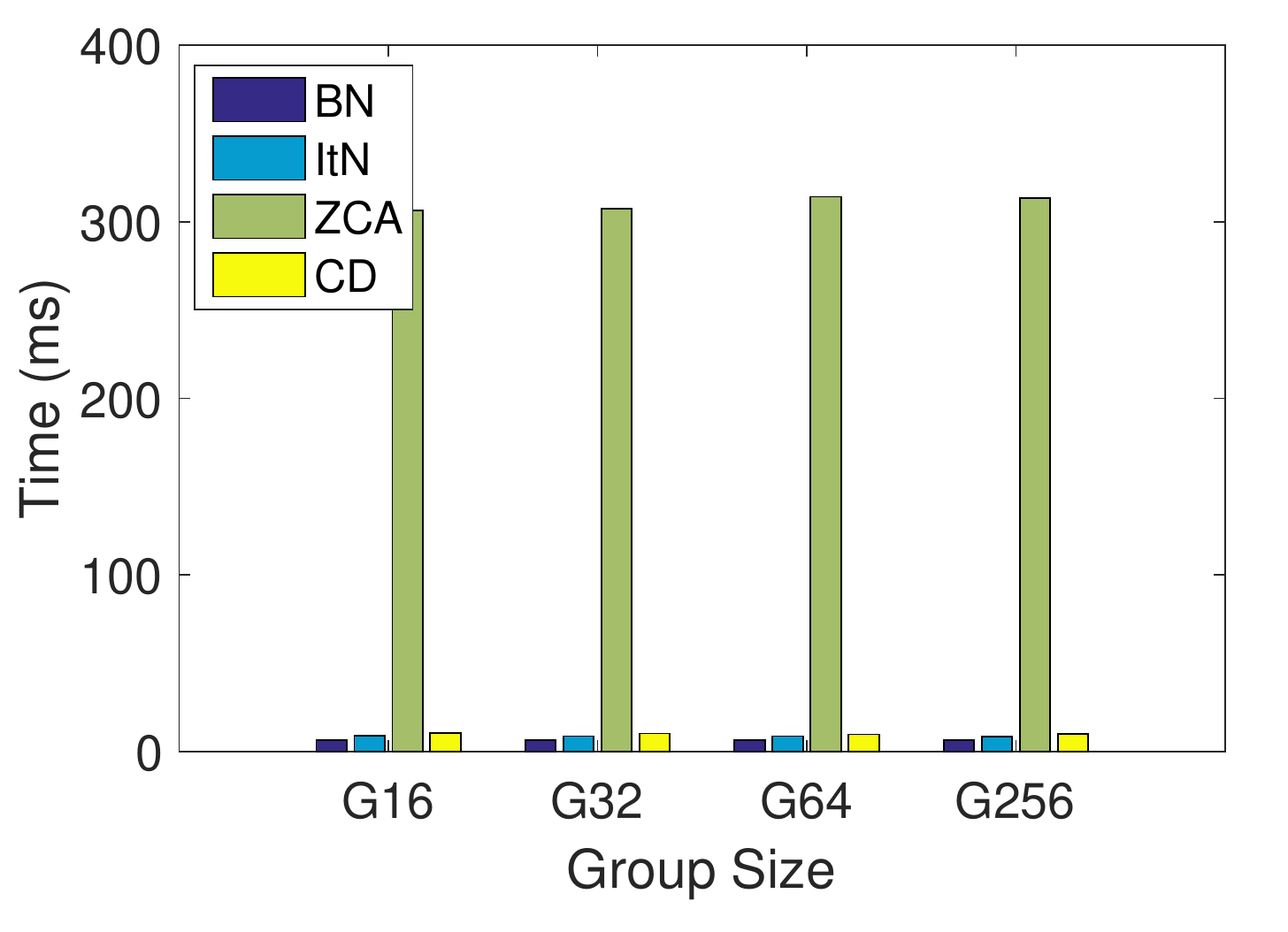}
		\end{minipage}
	}
	\hspace{0.15in}	\subfigure[CPU]{
		\begin{minipage}[c]{.44\linewidth}
			\centering
			\includegraphics[width=4.2cm]{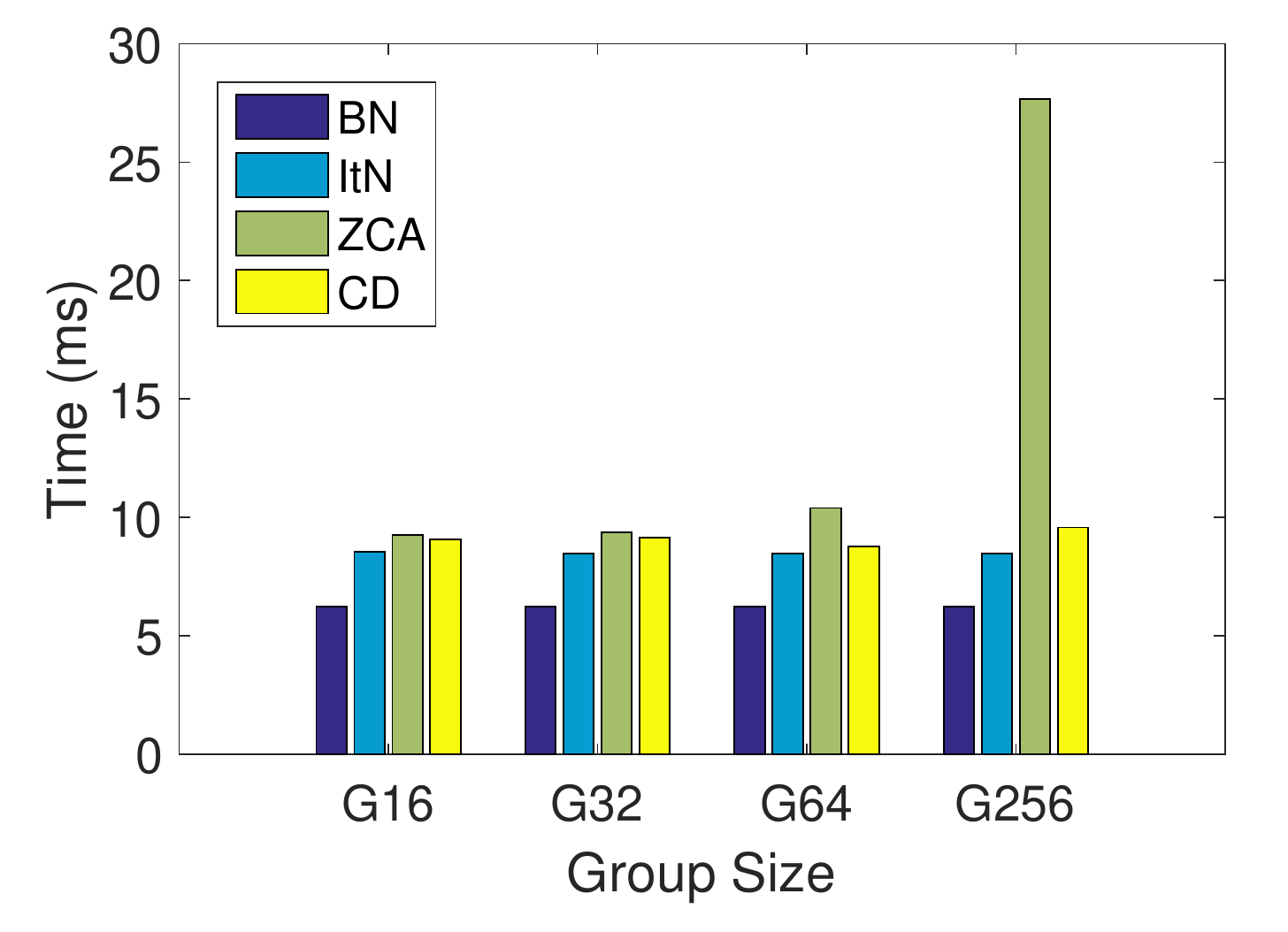}
		\end{minipage}
	}
	\caption{Time cost (ms/iteration) on GANs for stability experiments described in Section~\ref{sec:stability} of the paper. We compare the BW methods with different group sizes  running on  (a) a GPU and (b) a CPU. We evaluate the time cost of the generator where BW is applied, and the results are averaged over 100 iterations.}
	\label{fig:Time-GAN}
\end{figure}

\begin{table}[t]
	\centering
	\begin{footnotesize}
		\begin{tabular}{c|cccccc}
			\bottomrule[1pt]
			\multirow{2}{*}{Method}   
			&\multicolumn{6}{c}{Time (Group Size)} \\
			&16&32&64&128&256&512\\
			\hline
			BN   &0.049 &0.049 &0.049 & 0.049 &0.049 &0.049\\
			ItN  &0.624 &0.338 &0.192 & 0.13  &0.11  &0.106\\
			ZCA  &2.245 &1.536 &0.906 & 0.498 &0.505 &0.512\\
			CD   &9.8   &4.818 &2.391 & 1.29  &0.907 &0.793\\
			\toprule[1pt]
		\end{tabular}
		\caption{Time cost (s/iteration) on VGG for CIFAR-10 classification described in Section~\ref{sec:VGG} of the paper. We compare the BW methods with different group sizes, and the results are averaged over 100 iterations.}
		\label{tab:time vgg}
	\end{footnotesize}
\end{table}

\begin{table}[t]
	\centering
	\begin{footnotesize}
		\begin{tabular}{c|c}
			\bottomrule[1pt]
			Method   & Time  \\
			\hline
			Baselin (BN)          & 0.418 \\
			ItN-G256-ARC$_B$    & 0.464 \\
			ItN-G256-ARC$_C$ & 0.546 \\
			ItN-G64-ARC$_B$     & 0.522 \\
			ItN-G64-ARC$_C$  & 0.662 \\
			\toprule[1pt]
		\end{tabular}
		\caption{ Time cost (s/iteration) on ResNet-50 for ImageNet classification described in Section~\ref{exp_imagenet} of the paper. The results are averaged over 100 iterations.}
		\label{tab:time resnet 50}
	\end{footnotesize}
\end{table}

\begin{table}[t]
	\centering
	\begin{footnotesize}
		\begin{tabular}{c|cc|cc}
			\bottomrule[1pt]
			\multirow{2}{*}{Method}   
			&\multicolumn{2}{c|}{DCGAN}
			&\multicolumn{2}{c}{Resnet}\\
			&CPU&GPU&CPU&GPU\\
			\hline
			CD      &0.023 &0.019  &0.036 &0.038 \\
			ZCA-64  &0.018 &0.544  &0.036 &0.894  \\
			ItN     &0.017 &0.016  &0.034 &0.034 \\
			\toprule[1pt]
		\end{tabular}
		\caption{Time cost (s/iteration) on large GAN architecture described in Section~\ref{sec:larger} of the paper. We evaluate the time cost of the generator where BW is applied, and the results are averaged over 100 iterations. }
		\label{tab:time large gan}
	\end{footnotesize}
\end{table}

\begin{table*}[]
	\small
	\caption{DCGAN architectures in stability experiments.}
	\label{tab:stable dc}
	\centering
	\vspace{0.1cm}
	\subtable[Generator]{
		\begin{tabular}{c} 
			\hline 
			\hline
			$z \in \mathbb{R}^{128} \sim \mathcal{N}(0, I)$ \\
			\hline
			dense, $4\times 4\times 256$\\
			\hline
			$4\times 4$, stride=2 deconv. BN 256 ReLU\\
			\hline
			$4\times 4$, stride=2 deconv. BN 128 ReLU\\
			\hline
			$4\times 4$, stride=2 deconv. BN 64 ReLU\\
			\hline
			$3\times 3$, stride=1 conv. 3 Tanh\\
			\hline
			\hline
		\end{tabular}
	}
	\subtable[Discriminator]{
		\renewcommand\arraystretch{1.1}
		\begin{tabular}{c} 
			\hline 
			\hline
			RGB image $x \in \mathbb{R}^{32 \times 32 \times 3}$ \\
			\hline
			$3\times 3$, stride=1 conv 32 lReLU\\
			\hline
			$3\times 3$, stride=2 conv 64 lReLU\\
			$3\times 3$, stride=1 conv 64 lReLU\\
			\hline
			$3\times 3$, stride=2 conv 128 lReLU\\
			$3\times 3$, stride=1 conv 128 lReLU\\
			\hline
			$3\times 3$, stride=2 conv 256 lReLU\\
			$3\times 3$, stride=1 conv 256 lReLU\\
			\hline
			dense $\rightarrow$ 1 \\
			\hline
			\hline
		\end{tabular}
	}	
\end{table*}

\begin{table*}[]
	\small
	\caption{Large architecture of DCGAN.}
	\label{tab:large dc}
	\centering
	\vspace{0.1cm}
	\subtable[Generator]{
		\renewcommand\arraystretch{1.1}
		\begin{tabular}{c} 
			\hline 
			\hline
			$z \in \mathbb{R}^{128} \sim \mathcal{N}(0, I)$ \\
			\hline
			dense, $4\times 4\times 512$\\
			\hline
			$4\times 4$, stride=2 deconv. BN 512 ReLU\\
			\hline
			$4\times 4$, stride=2 deconv. BN 256 ReLU\\
			\hline
			$4\times 4$, stride=2 deconv. BN 128 ReLU\\
			\hline
			$3\times 3$, stride=1 conv. 3 Tanh\\
			\hline
			\hline
		\end{tabular}
		\vspace{0.1cm}
	}
	\subtable[Discriminator]{
		\renewcommand\arraystretch{1.1}
		\begin{tabular}{c} 
			\hline 
			\hline
			RGB image $x \in \mathbb{R}^{32 \times 32 \times 3}$ \\
			\hline
			$3\times 3$, stride=1 conv 64 lReLU\\
			\hline
			$4\times 4$, stride=2 conv 128 lReLU\\
			$3\times 3$, stride=1 conv 128 lReLU\\
			\hline
			$4\times 4$, stride=2 conv 256 lReLU\\
			$3\times 3$, stride=1 conv 256 lReLU\\
			\hline
			$4\times 4$, stride=2 conv 512 lReLU\\
			$3\times 3$, stride=1 conv 512 lReLU\\
			\hline
			dense $\rightarrow$ 1 \\
			\hline
			\hline
		\end{tabular}
		\vspace{0.1cm}
	}	
	\vspace{-0.1in}
\end{table*}

\begin{table*}[]
	\small
	\centering
	\caption{Large architecture of ResNet. The ResBlock  follows \cite{2019_ICLR_Siaroin}.}
	\label{tab:large res}
	\vspace{0.1cm}
	\subtable[Generator]{
		\centering
		\renewcommand\arraystretch{1.1}
		\begin{tabular}{c} 
			\hline 
			\hline
			$z \in \mathbb{R}^{128} \sim \mathcal{N}(0, I)$ \\
			\hline
			dense, $4\times 4\times 256$\\
			\hline
			ResBlock up 256\\
			\hline
			ResBlock up 256\\
			\hline
			ResBlock up 256\\
			\hline
			BN, ReLU, $3\times 3$ conv, 3 Tanh\\
			\hline
			\hline
		\end{tabular}
	}
	\subtable[Discriminator]{
		\centering
		\renewcommand\arraystretch{1.1}
		\begin{tabular}{c} 
			\hline 
			\hline
			RGB image $x \in \mathbb{R}^{32 \times 32 \times 3}$ \\
			\hline
			ResBlock down 128\\
			\hline
			ResBlock down 128\\
			\hline
			ResBlock 128\\
			\hline
			ResBlock 128\\
			\hline
			ReLU\\
			\hline
			Global sum pooling\\
			\hline
			dense $\rightarrow$ 1 \\
			\hline
			\hline
		\end{tabular}
	}	
\end{table*}

\subsubsection{GANs Using Tensorflow}
\label{sec:Gan_Tensor}
Figure \ref{fig:Time-GAN} shows the time costs of the GAN models  described in Section~\ref{sec:stability} of the paper. We perform  experiments with the BW operation running on both a GPU and a CPU, shown in Figure \ref{fig:Time-GAN} (a) and \ref{fig:Time-GAN} (b), respectively. We observe that ZCA whitening running on GPUs has a significantly more expensive computational cost than other methods, which is consistent with the observation in \cite{2019_ICLR_Siaroin}. We find this can be alleviated  when running ZCA whitening on CPUs. 

Table \ref{tab:time large gan} shows the time cost of the GAN models  described in Section~\ref{sec:larger} of the paper.



\end{document}